\PassOptionsToPackage{dvipsnames,table}{xcolor}
\RequirePackage{amsmath,amssymb}
\documentclass{bytedance}
\usepackage{xspace}
\usepackage[shortlabels,inline]{enumitem}
\usepackage{microtype}
\usepackage{mathtools}
\usepackage{array,tabularx,multirow,makecell,colortbl}
\usepackage{adjustbox}
\usepackage{listings}
\usepackage{needspace}
\pretocmd{\section}{\Needspace{8\baselineskip}}{}{}
\pretocmd{\subsection}{\Needspace{4\baselineskip}}{}{}
\newcommand{\beginsupplement}{%
  \clearpage\onecolumn\appendix
  \setcounter{page}{1}
  \renewcommand{\tablename}{Supplementary Table}
  \renewcommand{\figurename}{Supplementary Figure}
  \crefalias{table}{supplementarytable}
  \crefalias{figure}{supplementaryfigure}
  \raggedbottom
  \renewcommand{\thepage}{S\arabic{page}}
  \ifdefined\nolinenumbers\nolinenumbers\fi
  \begin{center}
    {\Large\sffamily\papertitle\par}
    \vspace{0.5em}{\large Supplementary Material\par}
  \end{center}\vspace{0.8em}
}

\usepackage{longtable,arydshln,pifont,comment,amsfonts}
\newcommand{\cmark}{\ding{51}}                        
\newcommand{\xmark}{\textcolor{black!50}{\ding{55}}}  
\newcommand{\hmark}{\ding{108}\kern-0.55em\ding{109}} 

\newcolumntype{?}{!{\vrule}}

\definecolor{myblue}{HTML}{7C86FF}
\definecolor{mygreen}{HTML}{C27AFF}

\lstdefinestyle{promptstyle}{
  basicstyle=\ttfamily\small,
  breaklines=true,
  breakatwhitespace=false,
  columns=fullflexible,
  keepspaces=true,
  showstringspaces=false,
  frame=single,
  xleftmargin=1em,
  xrightmargin=1em
}

\newcommand{\ML}[1]{\textcolor{purple}{#1}}

\lstdefinestyle{promptstyle}{basicstyle=\ttfamily\footnotesize,
  breaklines=true,breakatwhitespace=false,columns=fullflexible,
  keepspaces=true,showstringspaces=false,frame=single,
  rulecolor=\color{black!25},xleftmargin=0.8em,xrightmargin=0.8em}


\usepackage{titletoc}
\contentsmargin{2.7em}
\titlecontents{appsection}[1.8em]{\addvspace{4pt}\small\bfseries}
  {\contentslabel{1.8em}}{}{\titlerule*[0.65em]{.}\contentspage}
\titlecontents{appsubsection}[4.4em]{\small}
  {\contentslabel{2.6em}}{}{\titlerule*[0.65em]{.}\contentspage}

\AtEndPreamble{%
  \crefname{appendix}{Supplementary}{Supplementary}%
  \Crefname{appendix}{Supplementary}{Supplementary}%
  \crefname{supplementarytable}{Supplementary Table}{Supplementary Tables}%
  \Crefname{supplementarytable}{Supplementary Table}{Supplementary Tables}%
  \crefname{supplementaryfigure}{Supplementary Figure}{Supplementary Figures}%
  \Crefname{supplementaryfigure}{Supplementary Figure}{Supplementary Figures}%
}

\hypersetup{pdftitle={InteracVid: Building a Real Interactive Audio-Visual Response Dataset from Live-Chat Videos},
  pdfauthor={Chi Zhang, Junjie Huang, Shi Haoyang, Yueyi Liu, Zhaokun Yan, Yishu Yin, Zhaofeng Sun, Yuhang Wu, Jinzhao Li, Jiacheng Hua, Miao Liu}}
\newif\ifincludesupplement
\includesupplementtrue
\newcommand{\papertitle}{InteracVid: Building a Real Interactive Audio-Visual Response Dataset from Live-Chat Videos}
\title{InteracVid: Building a Real Interactive Audio-Visual Response Dataset from Live-Chat Videos}
\author[1,2]{Chi Zhang}
\author[1]{Junjie Huang}
\author[1,3]{Shi Haoyang}
\author[1,4]{Yueyi Liu}
\author[1]{Zhaokun Yan}
\author[1]{Yishu Yin}
\author[1]{Zhaofeng Sun}
\author[1]{Yuhang Wu}
\author[1,2]{Jinzhao Li}
\author[1]{Jiacheng Hua}
\author[1,\dagger]{Miao Liu}
\affiliation[1]{Tsinghua University}
\affiliation[2]{ByteDance}
\affiliation[3]{Fudan University}
\affiliation[4]{Xi'an Jiaotong University}
\contribution[\dagger]{Corresponding author}
\abstract{Audio-video generative models can synthesize high-fidelity synchronized content, but their supervision remains largely \emph{descriptive}: models render captions rather than respond to user interactions. We introduce \textbf{InteracVid}, \emph{the first open-source large-scale dataset for interactive audio-visual response generation}. Each sample couples preceding audio-visual context and a real or reconstructed user query with a real response. A metadata-aware pipeline extracts over \textbf{454K} context-query-response triplets from more than \textbf{59K} livestream videos, spanning conversation-centered, object-centric, procedural, embodied, and screen-based scenarios. A ten-rater study validates the causal plausibility, naturalness, and temporal completeness of interactions in both query branches. On a held-out benchmark of \textbf{100} genuine live-chat queries, fine-tuning on InteracVid improves both interaction planning and audio-video response generation. Independent human evaluation reproduces the system ranking and conclusions obtained with our automatic judge. These results highlight interaction-structured supervision as a foundation for grounded multimodal assistants, avatars, and embodied agents.
}
\correspondence{\email{imzc.2004@gmail.com}\quad\email{miaoliu@mail.tsinghua.edu.cn}}
\checkdata[Resources]{\href{https://interacvid.github.io}{Project Page}\quad
  \href{https://huggingface.co/datasets/Holomorphica/InteracVid-Parquet}{Dataset}\quad
  \href{https://github.com/InteracVid/InteracVid}{Code}}

\begin{document}
\maketitle
\begin{figure*}[t]
  \centering
  \captionsetup{font=small}
  \includegraphics[width=\textwidth]{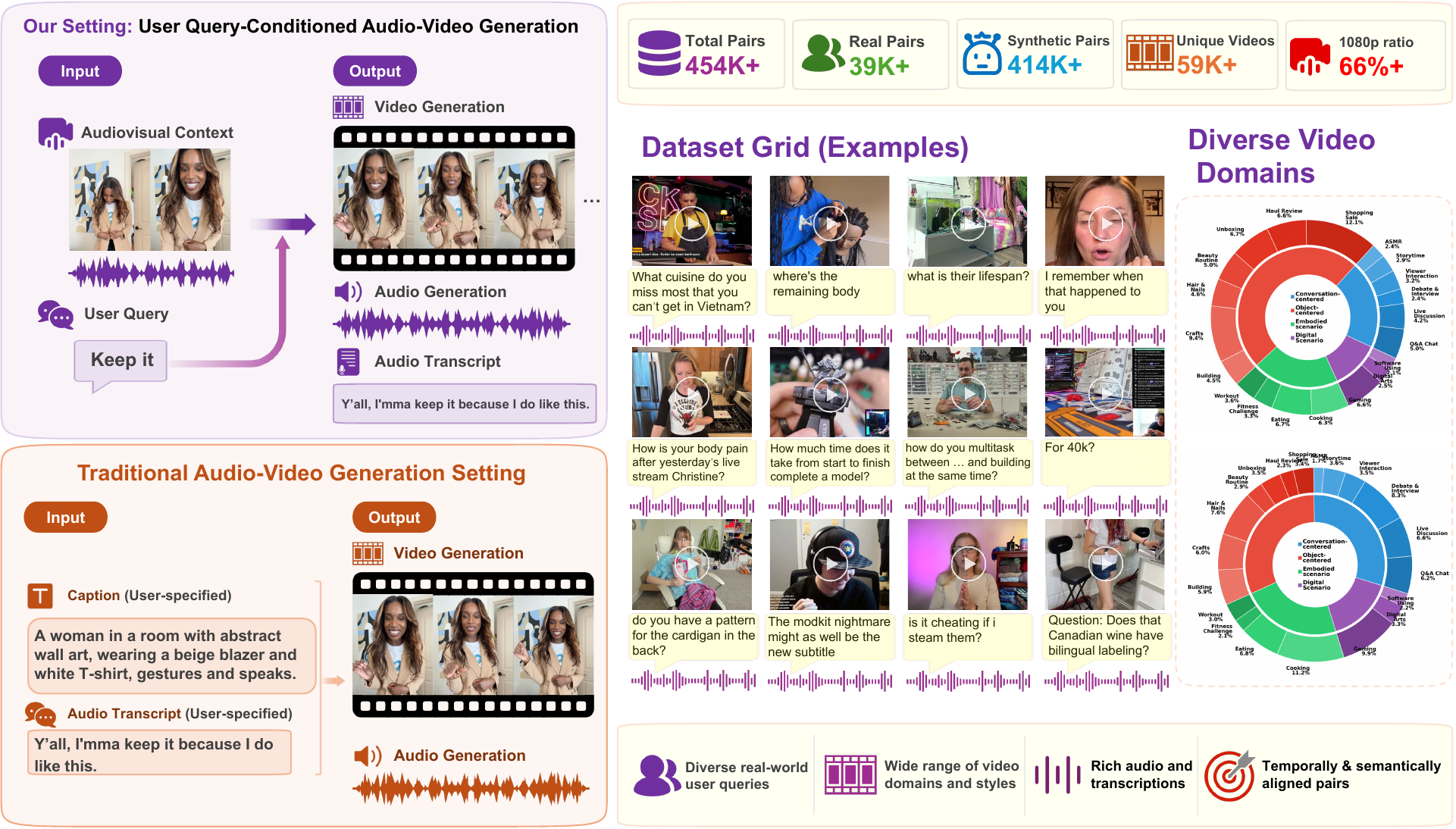}
  \caption{\emph{Overview of InteracVid.} Each sample pairs preceding audio-visual context and a real or reconstructed user query with a real response, spanning diverse livestream domains and interactions expressed through speech, facial reactions, and visible actions.}
  \label{fig:teaser}
\end{figure*}

\section{Introduction}
\label{sec:intro}
Advances in large language models~\cite{singh2025openai, comanici2025gemini, hurst2024gpt, brown2020language, achiam2023gpt} have made text the dominant medium for human--AI interaction. However, text-based interactions remain constrained by the response modality and struggle with expressiveness and multimodal grounding. As AI systems move from text chatbots toward multimodal assistants, avatars, and embodied agents, interaction should no longer be limited to generating the next utterance. A model should be able to generate an audio-visual response that is appropriate to the user's query and grounded in the visual context.

Recent audio-video generative models have substantially improved the fidelity and synchronization of multimodal synthesis. Existing systems can generate talking faces~\cite{xu2024vasa,tian2024emo, zhou2019talking, wang2019few, hong2023implicit,zhang2023sadtalker,guo2024liveportrait}, speech-driven video generation~\cite{yi2023generating, liu2024emage, xu2025hunyuanportrait, lin2025omnihuman,zhou2022responsive, liu2026elasticttt}, conversational avatars~\cite{low2025talkingmachines,wang2025omnitalker,pang2025mavid, gan2025omniavatar, chen2025hunyuanvideo,xu2024vasa,zhang2025voicebridge}, and text-conditioned audio-video clips~\cite{ruan2023mm, polyak2024movie, liu2025javisdit,wang2024av,huang2025jova,qiang2026mm,seedance2025seedance,seedance2026seedance}. However, most of these systems remain primarily \emph{descriptive}: they render multimodal content specified by input captions rather than producing responses caused by an external interaction. In real interactive scenarios, a user query does not merely describe the desired output. It acts as a stimulus that may lead to a spoken answer, a gesture, a facial reaction, an object manipulation, a screen-based action, or a visual demonstration.

To train an audio-video model that can respond interactively, we need examples that preserve the full interaction structure: the interaction context, the user stimulus, and the resulting audio-visual response. Existing audio-video datasets~\cite{chen2020vggsound, gemmeke2017audio, zhang2025speakervid} are usually prepared for \emph{descriptive} generation training, and do not identify the reactive video segments or provide a user-side query. Interactive talking-head and avatar datasets~\cite{cai2025towards,zhang2025speakervid} move closer to response generation, but they are mostly restricted to human speaking behavior. Importantly, current datasets do not provide sufficient supervision for learning query-triggered, context-grounded, interactive audio-visual responses.

\emph{Livestream videos provide a natural source for this missing supervision}. In livestreams, viewers engage in live-chats, sending time-synchronized comments, questions, and requests, while streamers react in speech and action within the continuous context. These live-chats act as external interaction triggers, and the following streamer behavior provides naturally grounded audio-visual responses. However, converting livestreams into training data is non-trivial: livestreams are long, noisy, sparsely interactive, and only a small fraction of comments are actually answered. Moreover, live-chat metadata is not always available, and response boundaries are often ambiguous.

Therefore, we introduce \textbf{InteracVid}, \emph{a large-scale dataset for interactive audio-visual response generation collected from live-chat videos}, as shown in \Cref{fig:teaser}. Each sample contains a preceding audio-visual context, a user query, a real response clip with synchronized video and audio, and an auxiliary response caption. Our construction uses two complementary branches. For videos with time-stamped live-chat metadata, we recover natural viewer--streamer interaction by aligning comments with subsequent streamer responses. For videos without such metadata, we identify reactive response moments in the transcript and reconstruct plausible triggering queries using the local audio-visual context. Importantly, in both branches the target response is always extracted from real livestream footage; the reconstructed subset is synthetic only in the query condition, never in the audio-visual target.

We analyze the dataset from statistical, semantic, acoustic, and geometric perspectives, showing broad coverage of video domains and high audio-visual quality, and we validate with a ten-rater human study that both genuine and reconstructed queries are causally aligned with responses whose boundaries match human annotation. Finally, we evaluate the dataset using a two-stage interactive generation pipeline: an interaction planner predicts an interaction-aware textual description from the context and query, and an audio-video co-generator synthesizes the final multimodal response. Fine-tuning on InteracVid improves both the upstream planner and the downstream audio-visual generator, demonstrating that the dataset provides effective supervision for interactive audio-visual response generation.

\section{Related Work}
\label{sec:related}

\begin{table}[t]
\centering
\small
\setlength{\tabcolsep}{5pt}
\renewcommand{\arraystretch}{1.10}
\caption{Comparison with prior livestream-based or interactive audio-visual datasets that pair an external stimulus with a (multi-modal) response.
\cmark{} = supported; \hmark{} = partial / limited; \xmark{} = not supported. \textbf{InteracVid} is the only entry that supports \emph{both} audio-visual input and output together with a preceding context, while covering diverse video domains.}
\label{tab:dataset_compare}
\begin{adjustbox}{max width=\linewidth}
\begin{tabular}{l c ? c c c ? c c ? c}
\toprule
                          &        & \multicolumn{3}{?c?}{\textsc{Stimulus / Input}} & \multicolumn{2}{c?}{\textsc{Interaction}} & \textsc{Coverage} \\
\cmidrule(lr){3-5} \cmidrule(lr){6-7} \cmidrule(lr){8-8}
\textbf{Dataset}          & \textbf{Year} & \makecell{Trigger} & \makecell{AV\\input} & \makecell{Pre-\\context} & \makecell{Agent\\response} & \makecell{AV\\output} & \makecell{Diverse\\activities} \\
\midrule
AVSD~\cite{AlAmri2019AudioVS}              & 2018 & \cmark & \cmark & \cmark & text & \xmark & \cmark \\
LiveBot~\cite{ma2019livebot}            & 2019 & \cmark & \hmark & \hmark & text & \xmark & \hmark \\
VideoIC~\cite{wang2020videoic}          & 2020 & \cmark & \hmark & \hmark & text & \xmark & \hmark \\
LiveChat~\cite{gao2023livechat}         & 2023 & \cmark & \xmark & \cmark & text & \xmark & \hmark \\
MovieLC~\cite{chen2023knowledge}          & 2024 & \cmark & \cmark & \hmark & text & \xmark & \hmark \\
AvaMERG~\cite{Zhang2025TowardsME}         & 2025 & \cmark & \cmark & \hmark & \cmark & \cmark & \hmark \\
SpeakerVid-5M~\cite{Zhang2025SpeakerVid5MAL} & 2025 & \hmark & \cmark & \hmark & \cmark & \cmark & \hmark \\
LiveStar~\cite{yang2025livestar}        & 2025 & \cmark & \cmark & \cmark & text & \xmark & \cmark \\
\midrule
\rowcolor{black!5}\textbf{InteracVid (ours)}         & 2026 & \cmark & \cmark & \cmark & \cmark & \cmark & \cmark \\
\bottomrule
\end{tabular}
\end{adjustbox}
\end{table}

\subsection{Audio-Visual Joint Generation and Interaction}

Prior work has explored talking-face synthesis, responsive listening generation, speech-driven gesture generation and text- or audio-driven avatar animation~\cite{zhou2022responsive,xu2024vasa,low2025talkingmachines,wang2025omnitalker}.
We focus on \textbf{audio-visual interactive systems}, which integrate multimodal understanding with synchronized speech, facial and body motion, and video generation~\cite{cai2025towards,pang2025mavid,xie2025x,jin2026sentiavatar,sun2025streamavatar}.
Recent systems~\cite{xie2025x,pang2025mavid} employ a thinker-actor framework, utilizing an Omni-VLM~\cite{xu2025qwen3} and a block-wise denoising diffusion head based on pretrained audio and video generation models~\cite{wan2025wan,du2024cosyvoice}. However, due to a lack of multimodal interactive data, the thinker and actor need to be trained separately on text conversations and descriptive videos, causing a data bottleneck in audio-visual appropriateness and expressiveness.

\subsection{Interactive Audio-Visual Datasets}
Interactive multimodal datasets include video-grounded dialogue, empathetic multimodal response generation and streaming video understanding~\cite{alamri2019audio,zhang2025speakervid,zhang2025towards,yang2025livestar}.
For livestream scenarios, prior work constructs datasets by pairing video context with time-synchronized live-chat comments for video understanding tasks~\cite{ma2019livebot, wang2020videoic, gao2023livechat, AlAmri2019AudioVS, chen2023knowledge}, or focuses on multi-party dialogue generation~\cite{zhang2025towards, zhang2025speakervid}.
However, these datasets are either limited to talking humans, treat audio-visual information as input rather than as a target, or lack the preceding audio-visual context necessary for interactive generation. In contrast, we use live-chat queries as \emph{external stimuli}, with the preceding and subsequent audio-visual content as context and target, enabling the study of audio-visual interaction in more general cases.

\section{Dataset Curation Pipeline}
\label{sec:curation}

\subsection{Problem Formulation}

We formalize the task of audio-visual interactive response generation. Given the preceding interaction context $\mathcal{C}$ and a user query $\mathcal{Q}$, the model generates an audiovisual response $\mathcal{Y} = \left(\mathcal{V}^{\mathrm{rsp}}, \mathcal{A}^{\mathrm{rsp}}\right)$, where $\mathcal{V}^{\mathrm{rsp}}$ and $\mathcal{A}^{\mathrm{rsp}}$ denote the response video and audio, respectively. Unlike text-only dialogue generation or talking-head generation, our target response couples speech with synchronized facial and bodily reactions, object interactions, and scene context.

\subsection{Data Source and Pipeline Overview}

Prior audiovisual interaction pipelines~\cite{pang2025mavid,xie2025x} train text response generation and audiovisual response generation on separate data sources. To address this limitation and to develop end-to-end models, we curate a large-scale dataset where each sample is represented as $\mathcal{X} = \left(\mathcal{C}, \mathcal{Q}, \mathcal{Y}, \mathcal{T}\right)$, and $\mathcal{T}$ denotes an auxiliary audiovisual caption describing the response clip. The key property of this data schema is alignment in both semantic and low-level properties. Most importantly, the query $\mathcal{Q}$ should serve as the trigger or semantic cause of the response $\mathcal{Y}$. The response clip should also preserve temporal coherence and scene and speaker consistency with the preceding context $\mathcal{C}$.

\begin{figure*}[t]
    \centering
    \includegraphics[width=\linewidth]{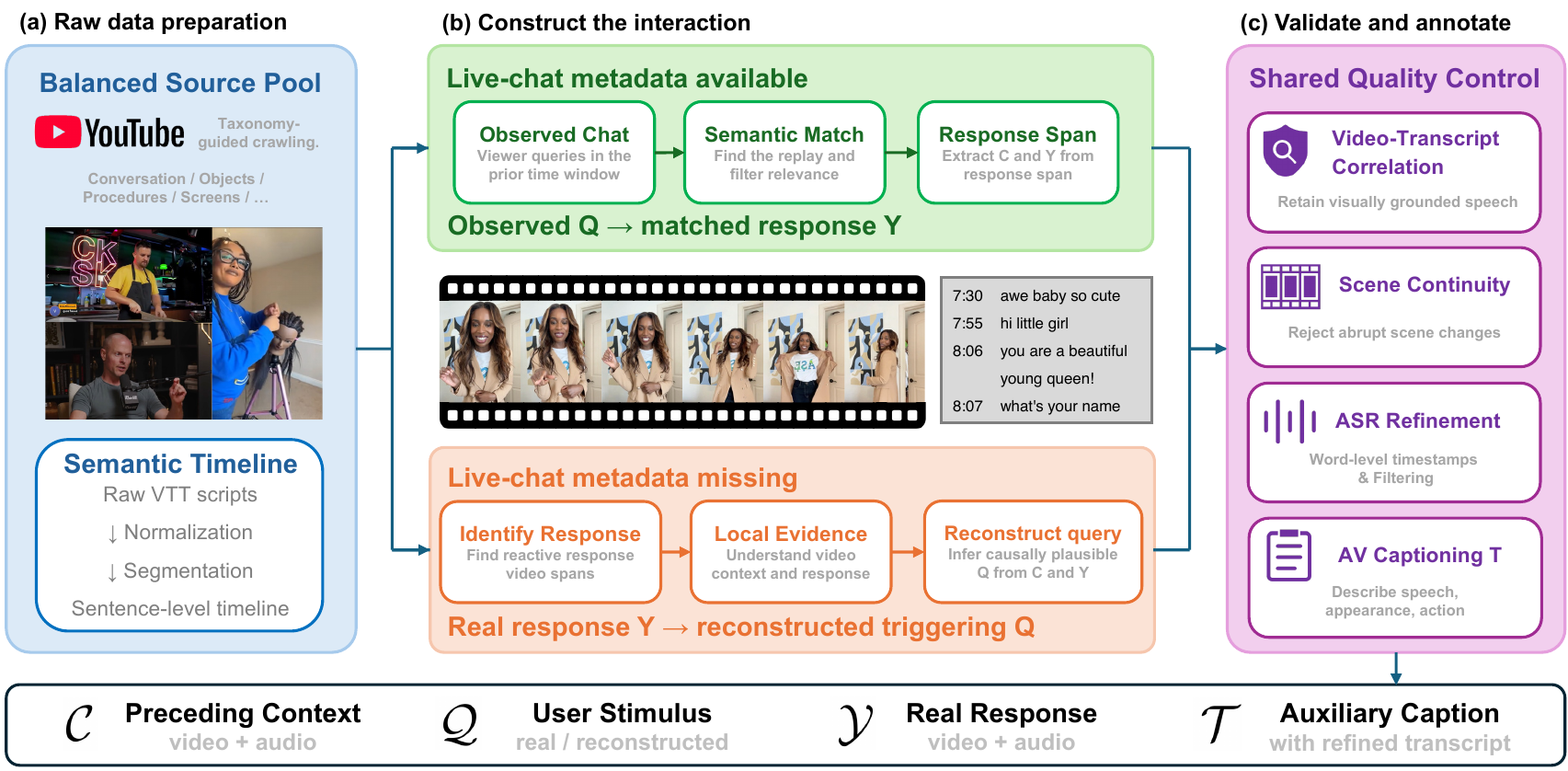}
    \caption{\emph{Dataset curation pipeline.} Taxonomy-guided crawling and transcript preparation feed two metadata-aware interaction branches, followed by shared multimodal filtering, transcript refinement, and auxiliary audio-visual captioning of the real response clips.}
    \label{fig:pipeline}
\end{figure*}

\emph{Livestream videos with live-chats provide an ideal source for curating such data due to their realistic interaction patterns}. Livestreams capture spontaneous human reactions, conversational dynamics, and diverse visual interactions. The accompanying live-chat comments offer rich user queries that are temporally aligned with the video content and responded to by the streamer, naturally forming query--response pairs grounded in audiovisual context. Two properties of the raw source shape the design of our pipeline: naively crawling livestreams yields unbalanced video domains, and time-stamped live-chat metadata is preserved by some YouTube livestream replays but absent from many Video-on-Demands (VODs) and stream recordings, which carry only video, audio, and subtitles. We therefore adopt a taxonomy-guided crawling strategy to obtain raw videos, and design a metadata-aware curation pipeline that converts long livestream videos into temporally localized interactive video samples. The overall pipeline is shown in \Cref{fig:pipeline}.

\subsection{Metadata-Aware Interaction Construction}
\label{sec:interaction_reconstruction}
We collect candidate livestream replays and VOD videos from YouTube. To improve coverage, we use a taxonomy-guided crawling strategy over conversation-centered, lifestyle/object-centric, and procedural/embodied scenarios, combining content keywords with livestream-oriented terms such as \textit{live}, \textit{livestream}, and \textit{VOD}. We retain a balanced candidate pool across subclasses to reduce genre dominance. The complete taxonomy and keyword lists are provided in \Cref{app:crawling}.

We then partition videos according to the availability of time-stamped live-chat or live-chat replay metadata. Videos with such metadata are used to construct the real-query subset, where $\mathcal{Q}$ is naturally observed. Videos without metadata are used for the reconstructed-query subset, where only the missing user utterance $\mathcal{Q}$ is synthesized. Importantly, in both subsets the target response $\mathcal{Y}$ is always extracted from real livestream footage. Thus, the reconstructed subset remains grounded in real audiovisual reactions: it is synthetic only with respect to the input query condition, not the output response. \Cref{sec:human_validation} quantifies, with human raters, how faithful this reconstruction is.

Both branches share a sentence-level semantic timeline that represents the discourse structure of the livestream. Rather than treating each short clip as an isolated unit, we first recover the stream-level semantic flow, since the interaction role of an utterance is often determined by its surrounding context. A sentence may be a proactive monologue, a direct response, a continuation, or a topic shift, and these roles are difficult to distinguish from the local clip alone. We therefore construct a temporally ordered transcript representation by collecting subtitle tracks, splitting raw Video Text Track (VTT) files into manageable chunks, and normalizing them with an instruction-tuned language model into complete timestamped sentences:
\begin{align}
\mathcal{S}=\{(s_i,t_i)\}_{i=1}^{N},
\end{align}
where $s_i$ is a complete transcript sentence and $t_i$ is its timestamp. This timeline serves as a searchable semantic backbone for locating interaction events in long livestreams, while preserving the global semantic flow needed to judge the role of each local sentence.

For the real-query branch, we recover naturally occurring viewer--streamer interactions. For each transcript sentence $(s_i,t_i)$, we collect live-chat comments sent within a preceding window:
\begin{align}
\mathcal{D}_i=\{d_j \mid t_i-t_\text{window}\leq \tau_j\leq t_i\},
\end{align}
where $d_j$ is a live-chat comment and $\tau_j$ is its timestamp. Since not all live chats are answered by the streamer, we perform semantic correspondence detection using an LLM over the sentence-level timeline, retaining only live-chat--sentence pairs that exhibit a plausible query--response relation. Once a match is found, we append the next five transcript sentences and select the subset that continues the same response. This yields a response span and its corresponding real audiovisual segment $\mathcal{Y}$.

For the reconstructed-query branch, we perform the inverse construction. We first use the global semantic timeline to classify transcript sentences into discourse roles: proactive monologue, reactive response, or continuation. This classification is determined by both the local sentence and its role within the stream-level semantic flow. For example, a sentence may be identified as a reactive response because it answers an implicit question raised by the preceding context, explains a visible event, corrects a previous action, or continues a response that began earlier. Reactive responses are then treated as real livestream moments likely triggered by an external stimulus. For each such span, a vision-language model observes the local video context together with the response transcript, and reconstructs a plausible user query that could have triggered the response.

\begin{figure*}[!t]
    \centering
    \captionsetup[subfigure]{font=footnotesize}
    \begin{subfigure}[t]{0.32\textwidth}
        \centering
        \includegraphics[width=\linewidth]{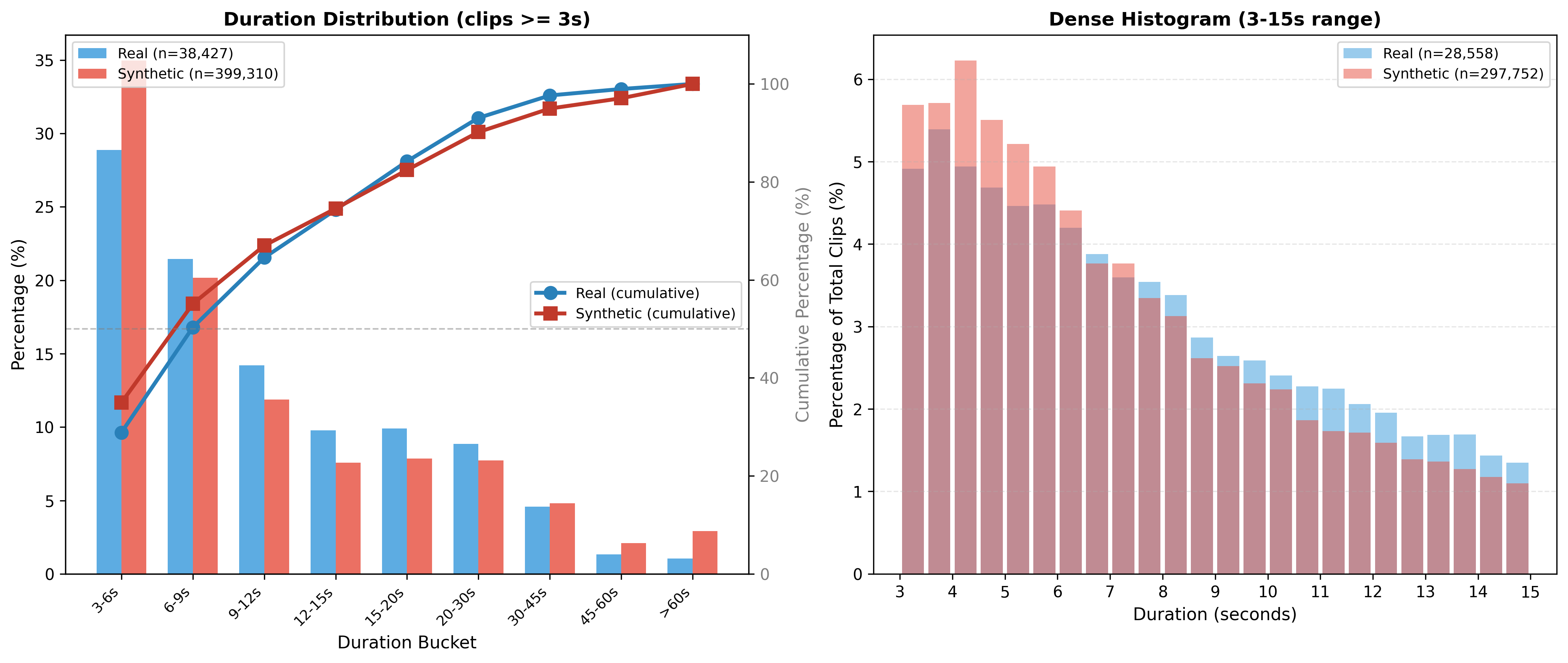}
        \caption{Video duration, both branches.}
        \label{fig:duration}
    \end{subfigure}
    \hfill
    \begin{subfigure}[t]{0.32\textwidth}
        \centering
        \includegraphics[width=\linewidth]{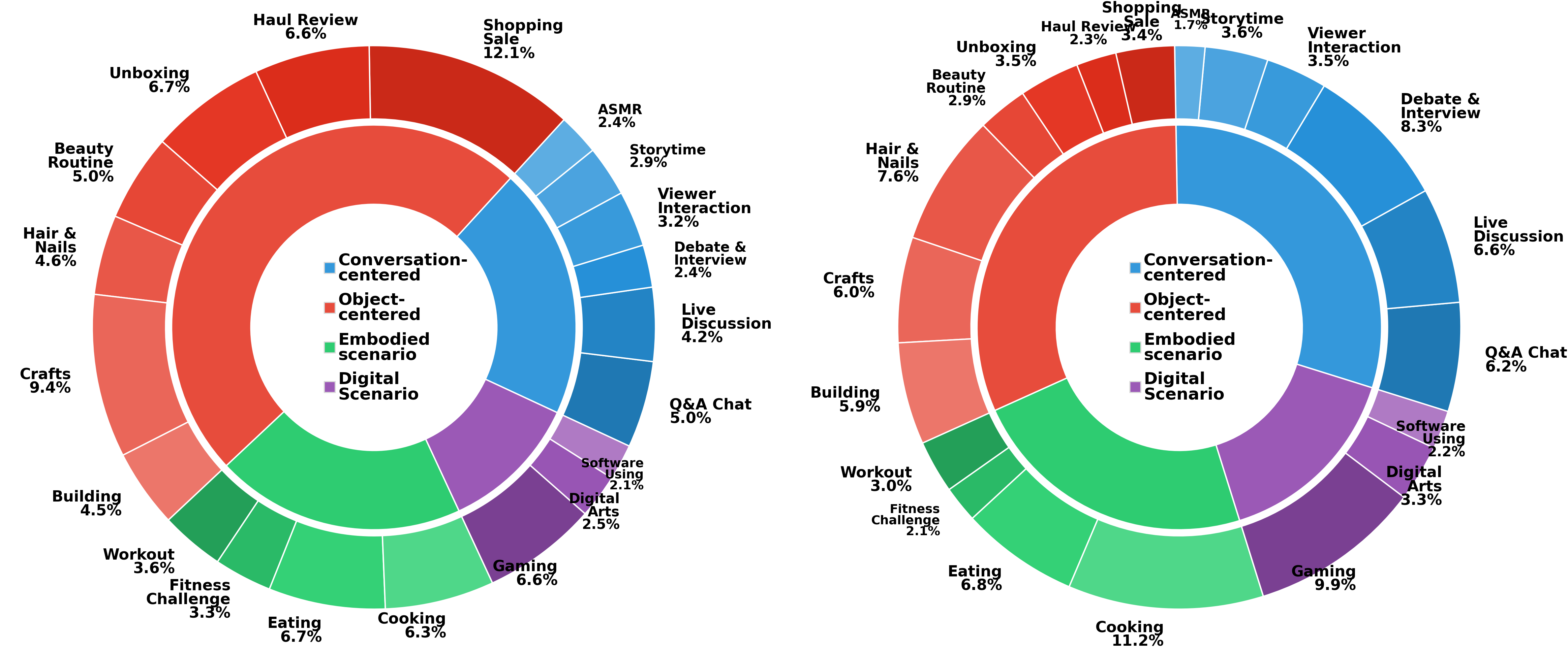}
        \caption{Domains: real (left), reconstructed (right).}
        \label{fig:class_pie}
    \end{subfigure}
    \hfill
    \begin{subfigure}[t]{0.32\textwidth}
        \centering
        \includegraphics[width=\linewidth]{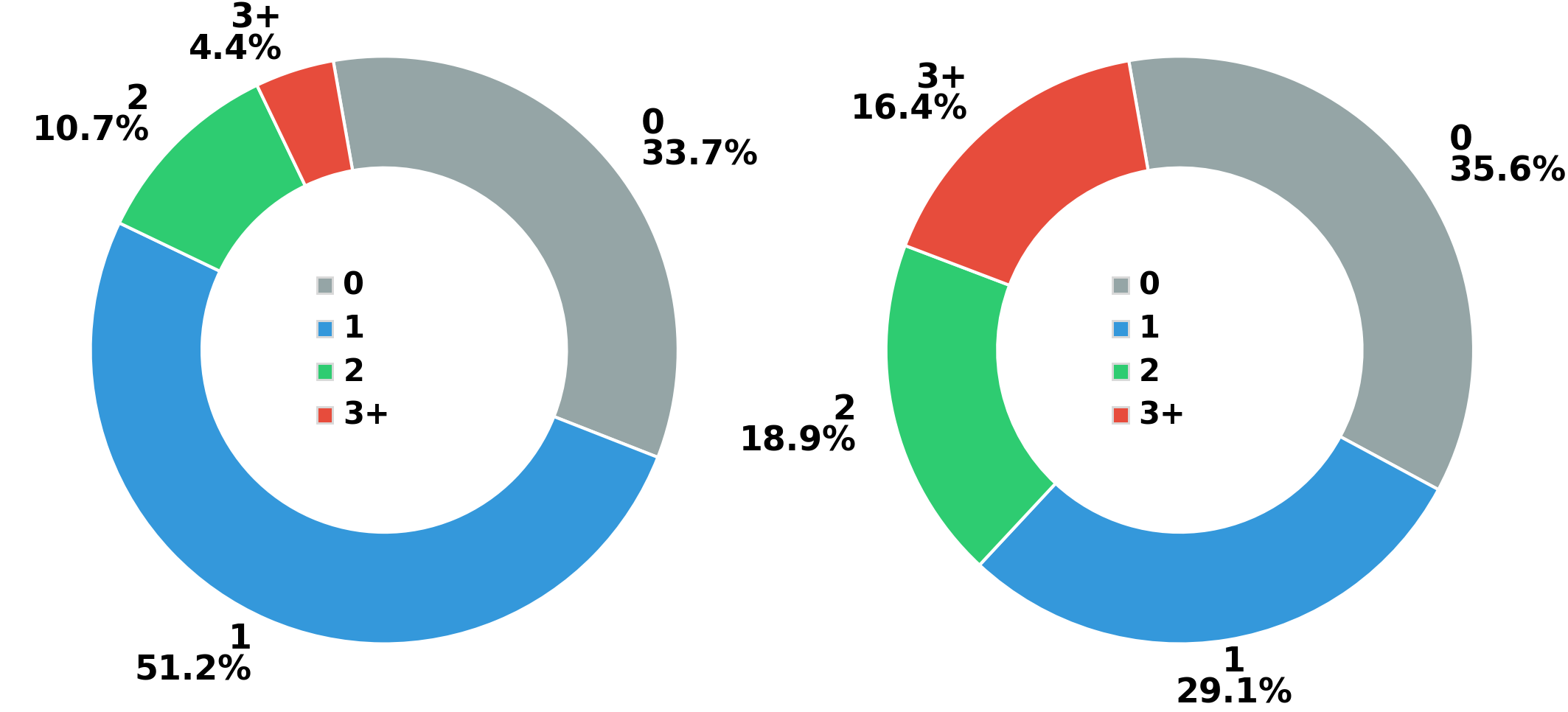}
        \caption{Human-count distributions in the main window for real (left) and reconstructed (right) queries.}
        \label{fig:human_count}
    \end{subfigure}
    \vspace{0.7em}
    \begin{subfigure}[t]{0.18\textwidth}
        \centering
        \includegraphics[width=\linewidth]{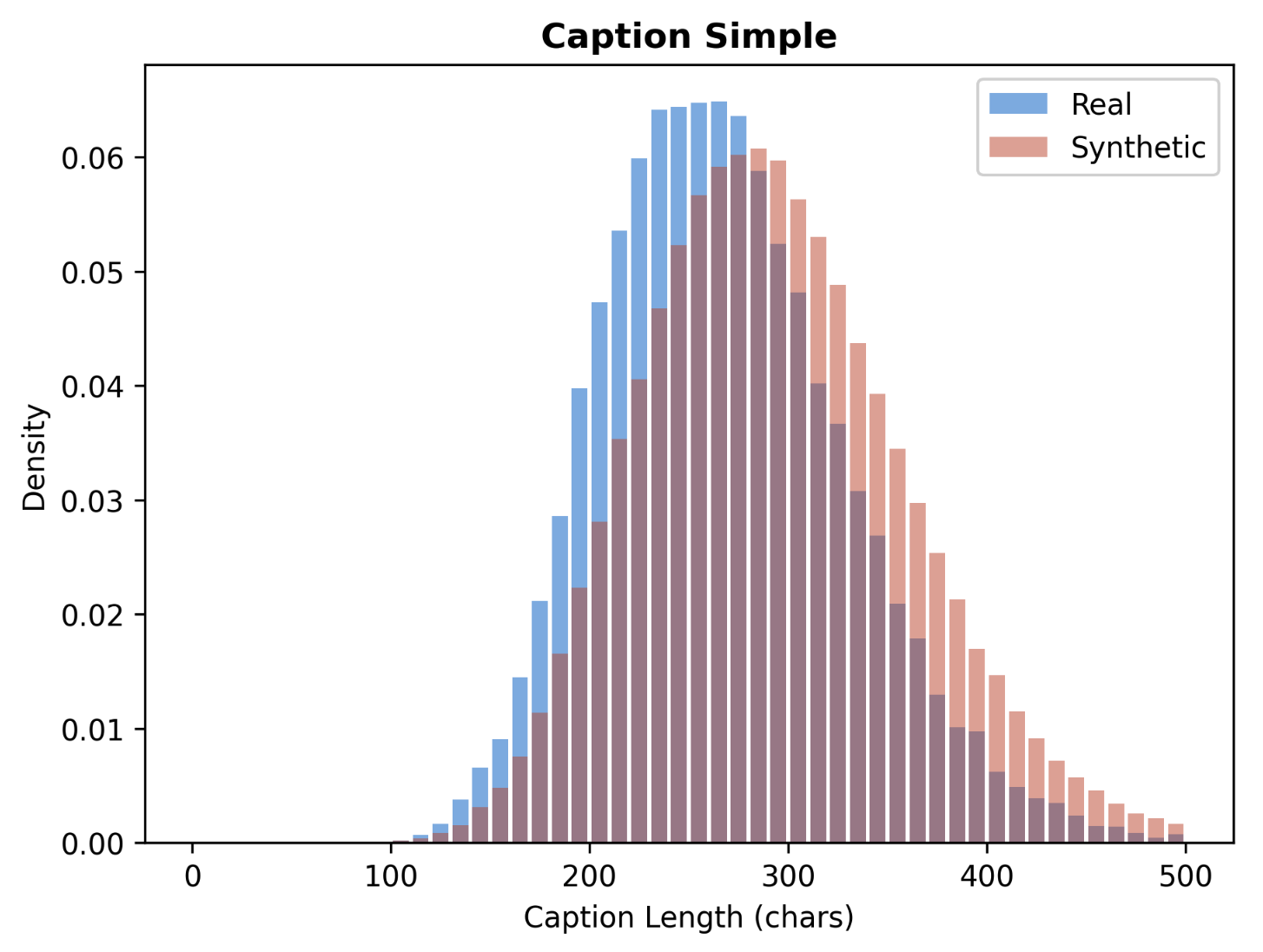}
        \caption{Caption lengths.}
        \label{fig:caption_length}
    \end{subfigure}
    \hfill
    \begin{subfigure}[t]{0.22\textwidth}
        \centering
        \includegraphics[width=\linewidth]{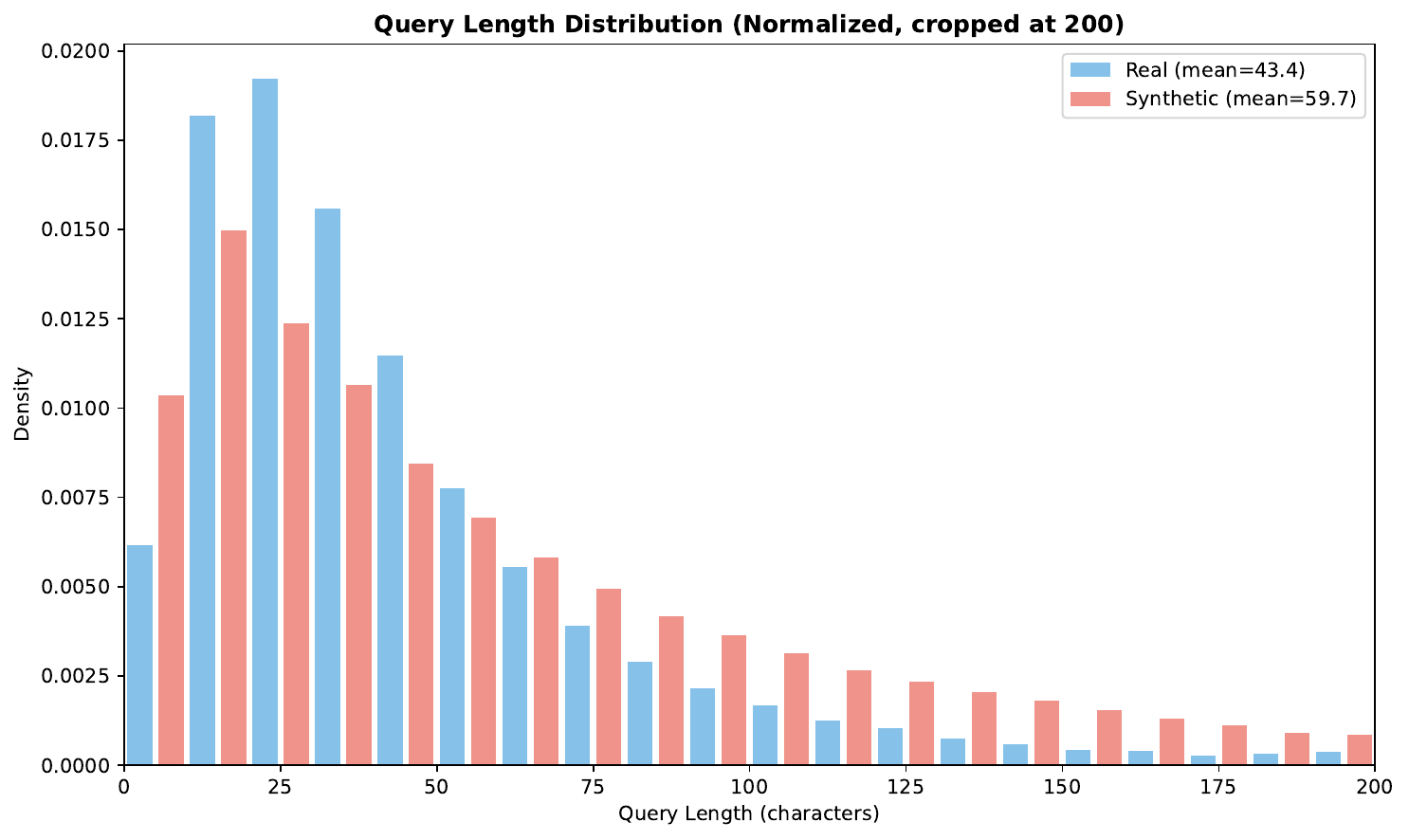}
        \caption{Query lengths.}
        \label{fig:query_length}
    \end{subfigure}
    \hfill
    \begin{subfigure}[t]{0.56\textwidth}
        \centering
        \includegraphics[width=\linewidth]{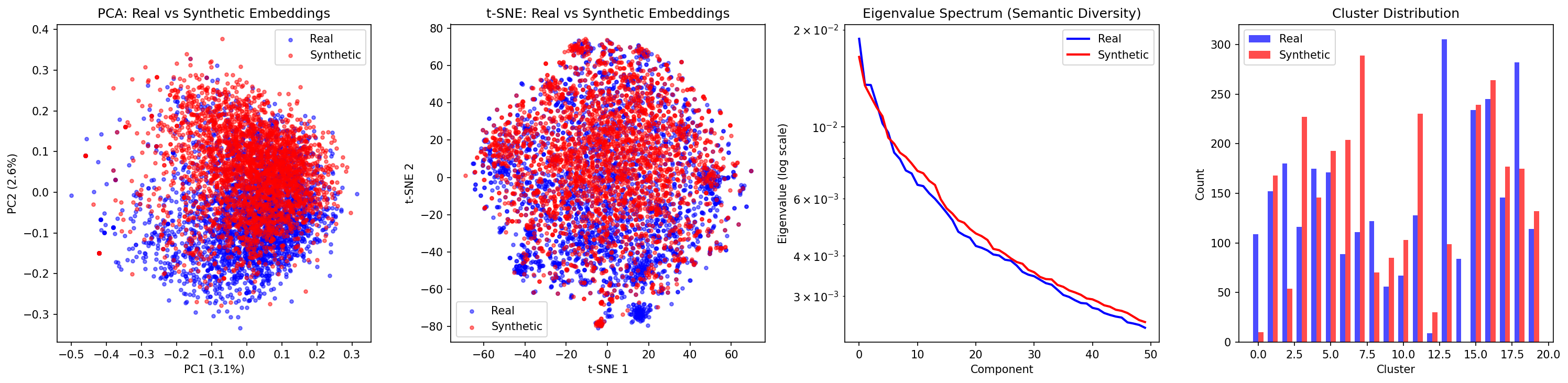}
        \caption{Semantic statistics: PCA, t-SNE, eigenvalues, clusters.}
        \label{fig:semantic_stats}
    \end{subfigure}
    \caption{\emph{Dataset statistics and query semantics.} We compare response duration, domain coverage, visible human count, caption and query lengths, and query embedding distributions (PCA, t-SNE, eigenspectra, and clusters) between the real and reconstructed branches.}
    \label{fig:dataset_statistics}
\end{figure*}

\subsection{Multimodal Quality Filtering and Captioning}

We apply rigorous multimodal quality control to ensure that each sample supports audiovisual reactive response generation rather than text-only dialogue modeling. We first verify the alignment between the response transcript and the visual content using a vision-language model, retaining only samples where the speech is directly grounded in the corresponding video segment. We also use scene-change detection to remove abrupt transitions that disrupt temporal coherence between the context $\mathcal{C}$ and response $\mathcal{Y}$.

To improve transcript quality, we rerun Automatic Speech Recognition (ASR) with Seed-ASR~\cite{bai2024seedasr} on the extracted response clip to obtain more accurate local transcriptions, and discard samples whose ASR outputs substantially deviate from the original response. The refined ASR output provides word-level timestamps, which allows the response clip to be recut to any duration required by a downstream generator. Finally, we generate an auxiliary audiovisual caption $\mathcal{T}$ by combining the refined ASR transcript with visual observations from a vision-language model. The resulting caption provides a structured description of the response clip that supports training, filtering, and analysis.

Because the response localization and the relevance filter both operate on speech, we additionally audit the acoustic composition of the corpus with an audio language model and release the resulting per-clip audio-pattern annotations, so that users can filter or stratify by the presence of background music, sound effects, and ambient noise. The outcome of the audit is reported in \Cref{app:audio_audit}.

\section{Dataset Characteristics}

\label{sec:statistics}

\begin{table}[t]
\centering
\small
\setlength{\tabcolsep}{6pt}
\renewcommand{\arraystretch}{1.05}
\caption{\emph{Human validation.} Mean 1--5 ratings and start/end span agreement for real and reconstructed queries.}
\label{tab:human_validation}
\begin{adjustbox}{max width=\linewidth}
\begin{tabular}{l ccc ? cc}
\toprule
 & \multicolumn{3}{c?}{\textsc{Rating}\,(1--5)} & \multicolumn{2}{c}{\textsc{Span agr.}} \\
\cmidrule(lr){2-4} \cmidrule(lr){5-6}
Query branch & Causality & Naturalness & Completeness & Start & End \\
\midrule
Real          & 4.09 & 4.17 & 4.14 & 93\% & 88\% \\
Reconstructed & \textbf{4.20} & \textbf{4.30} & \textbf{4.17} & \textbf{95\%} & \textbf{90\%} \\
\bottomrule
\end{tabular}
\end{adjustbox}
\end{table}

\begin{table*}[!t]
\centering
\small
\caption{\emph{Interactive response quality.} Left: VLM means $\pm$ standard errors on 100 held-out examples. Right: human means on the human-evaluation subset. OVRL averages the relevance, appropriateness, and expressiveness scores for each configuration.}
\label{tab:main}
\begin{adjustbox}{max width=\linewidth}
\setlength{\tabcolsep}{4pt}
\renewcommand{\arraystretch}{1.05}
\begin{tabular}{l c c c c ? c c c c c c}
\toprule
 & \multicolumn{4}{c?}{\cellcolor{myblue!30}\textsc{VLM judge}} & \multicolumn{6}{c}{\cellcolor{mygreen!30}\textsc{Human raters}} \\
\cmidrule(lr){2-5} \cmidrule(lr){6-11}
Planner + Generator
 & \textsc{Relev.}\,$\uparrow$ & \textsc{Approp.}\,$\uparrow$ & \textsc{Expr.}\,$\uparrow$ & \textsc{OVRL}\,$\uparrow$
 & \textsc{Relev.}\,$\uparrow$ & \textsc{Approp.}\,$\uparrow$ & \textsc{Expr.}\,$\uparrow$ & \textsc{Vid.}\,$\uparrow$ & \textsc{Aud.}\,$\uparrow$ & \textsc{OVRL}\,$\uparrow$ \\
\midrule
Oracle + MOVA \emph{(pt.)}
 & 4.89\,$\pm$\,0.042 & \textbf{4.61\,$\pm$\,0.108} & 3.15\,$\pm$\,0.10 & 4.22
 & 4.33 & 3.89 & 3.72 & 3.17 & 3.61 & 3.98 \\
Oracle + MOVA \emph{(ft.)}
 & \textbf{4.96\,$\pm$\,0.03} & 4.54\,$\pm$\,0.083 & \textbf{3.41\,$\pm$\,0.09} & \textbf{4.31}
 & \textbf{4.41} & \textbf{3.98} & \textbf{3.96} & \textbf{4.00} & \textbf{3.98} & \textbf{4.12} \\
\hspace{1em}$\Delta$\;\textcolor{black!50}{(+Gen. ft.)}
 & \textcolor{teal}{+0.07} & \textcolor{black!45}{$-$0.07} & \textcolor{teal}{+0.26} & \textcolor{teal}{+0.09}
 & \textcolor{teal}{+0.08} & \textcolor{teal}{+0.09} & \textcolor{teal}{+0.24} & \textcolor{teal}{+0.83} & \textcolor{teal}{+0.37} & \textcolor{teal}{+0.14} \\
\midrule
Qwen3.5 \emph{(pt.)} + MOVA \emph{(pt.)}
 & 1.85\,$\pm$\,0.19 & 1.65\,$\pm$\,0.17 & 1.73\,$\pm$\,0.09 & 1.74
 & 2.62 & 1.92 & 2.22 & 1.93 & 2.48 & 2.25 \\
Qwen3.5 \emph{(pt.)} + MOVA \emph{(ft.)}
 & 3.60\,$\pm$\,0.16 & 2.99\,$\pm$\,0.19 & 2.88\,$\pm$\,0.09 & 3.16
 & 3.55 & 3.14 & 3.17 & 3.53 & 3.79 & 3.29 \\
Qwen3.5 \emph{(ft.)} + MOVA \emph{(ft.)}
 & \textbf{3.91\,$\pm$\,0.16} & \textbf{3.63\,$\pm$\,0.19} & \textbf{3.25\,$\pm$\,0.09} & \textbf{3.60}
 & \textbf{4.00} & \textbf{3.59} & \textbf{3.63} & \textbf{3.81} & \textbf{3.96} & \textbf{3.74} \\
\hspace{1em}$\Delta$\;\textcolor{black!50}{(+Gen. ft.)}
 & \textcolor{teal}{+1.75} & \textcolor{teal}{+1.34} & \textcolor{teal}{+1.15} & \textcolor{teal}{+1.41}
 & \textcolor{teal}{+0.93} & \textcolor{teal}{+1.22} & \textcolor{teal}{+0.95} & \textcolor{teal}{+1.60} & \textcolor{teal}{+1.31} & \textcolor{teal}{+1.04} \\
\hspace{1em}$\Delta$\;\textcolor{black!50}{(+Plan. ft.)}
 & \textcolor{teal}{+0.31} & \textcolor{teal}{+0.64} & \textcolor{teal}{+0.37} & \textcolor{teal}{+0.44}
 & \textcolor{teal}{+0.45} & \textcolor{teal}{+0.45} & \textcolor{teal}{+0.46} & \textcolor{teal}{+0.28} & \textcolor{teal}{+0.17} & \textcolor{teal}{+0.45} \\
\midrule
InternVL3.5 \emph{(pt.)} + MOVA \emph{(pt.)}
 & 1.79\,$\pm$\,0.18 & 1.43\,$\pm$\,0.121 & 1.56\,$\pm$\,0.09 & 1.59
 & 2.04 & 1.63 & 1.48 & 1.37 & 1.81 & 1.72 \\
InternVL3.5 \emph{(pt.)} + MOVA \emph{(ft.)}
 & 3.60\,$\pm$\,0.17 & 3.24\,$\pm$\,0.18 & 2.79\,$\pm$\,0.09 & 3.21
 & 3.86 & 3.08 & 3.26 & \textbf{3.86} & \textbf{3.84} & 3.40 \\
InternVL3.5 \emph{(ft.)} + MOVA \emph{(ft.)}
 & \textbf{3.89\,$\pm$\,0.10} & \textbf{3.36\,$\pm$\,0.16} & \textbf{2.89\,$\pm$\,0.09} & \textbf{3.38}
 & \textbf{3.96} & \textbf{3.48} & \textbf{3.63} & 3.63 & 3.78 & \textbf{3.69} \\
\hspace{1em}$\Delta$\;\textcolor{black!50}{(+Gen. ft.)}
 & \textcolor{teal}{+1.81} & \textcolor{teal}{+1.81} & \textcolor{teal}{+1.23} & \textcolor{teal}{+1.62}
 & \textcolor{teal}{+1.82} & \textcolor{teal}{+1.45} & \textcolor{teal}{+1.78} & \textcolor{teal}{+2.49} & \textcolor{teal}{+2.03} & \textcolor{teal}{+1.68} \\
\hspace{1em}$\Delta$\;\textcolor{black!50}{(+Plan. ft.)}
 & \textcolor{teal}{+0.29} & \textcolor{teal}{+0.12} & \textcolor{teal}{+0.10} & \textcolor{teal}{+0.17}
 & \textcolor{teal}{+0.10} & \textcolor{teal}{+0.40} & \textcolor{teal}{+0.37} & \textcolor{black!45}{$-$0.23} & \textcolor{black!45}{$-$0.06} & \textcolor{teal}{+0.29} \\
\bottomrule
\end{tabular}

\end{adjustbox}
\end{table*}

\subsection{Scale and Basic Statistics}

We crawled over \textbf{160K} livestreams totaling over \textbf{185K} hours. Curation yields over \textbf{454K} clips from \textbf{59K} livestreams: \textbf{39K} clips totaling \textbf{76} hours in the real-query branch and \textbf{414K} clips totaling \textbf{903} hours in the reconstructed-query branch. The ratio between raw and retained duration shows how sparsely genuine interaction occurs in livestreams, and is precisely what the two complementary branches are designed to overcome.

We report the basic dataset statistics in \Cref{fig:dataset_statistics}. As shown in \Cref{fig:class_pie}, our dataset exhibits diverse and relatively balanced scenario coverage, including conversation-centered videos, object-centered manipulation videos, procedural and embodied scenarios, and digital scenarios. The category distribution shows mild differences between the two branches, but both maintain broad domain coverage.

We further analyze the distributions of video duration, live-chat query length, caption length, and the number of visible humans. The response videos span a wide temporal range, from approximately 3 seconds to over 60 seconds, reflecting different levels of streamer elaboration in response to external queries. Since the videos are accompanied by sentence-level timestamps and word-level ASR, they can also be flexibly segmented into shorter clips compatible with the input duration constraints of existing video generation models. The query length distributions of the two branches are broadly similar, with both peaking at around 20 characters, while reconstructed queries exhibit a slightly heavier long tail. In terms of human presence, most samples contain zero or one visible human in the main window; livestreams with a small face-cam overlay are counted as zero-human main-window cases. A smaller proportion of samples involve multi-human scenarios.

\subsection{Semantic Analysis of Reconstructed Queries}

To validate the quality of reconstructed live-chat queries, we conduct a semantic distribution analysis between the two subsets. Specifically, we extract query-level semantic embeddings using a Sentence-BERT model~\cite{Reimers2019SentenceBERTSE, devlin2019bert} following~\cite{gao2023livechat}, and visualize the resulting latent distributions with different visualization techniques in \Cref{fig:semantic_stats}. The two subsets show substantial overlap in the semantic space, indicating that reconstructed queries largely preserve the semantic structure of real live-chat. Meanwhile, as shown in the PCA~\cite{mackiewicz1993principal} and t-SNE~\cite{van2008visualizing} figures, real queries cover a broader peripheral region despite having fewer samples, reflecting the wider spread of intents in spontaneous live chat. This is consistent with the role of the reconstructed branch as a controlled expansion of real interaction patterns.

\subsection{Human Validation of Interaction Quality}
\label{sec:human_validation}

To complement the distributional analysis, ten raters, blind to query source, scored pairs sampled across both branches and all four domains for \emph{causality}, \emph{naturalness}, and \emph{completeness} on a 1--5 scale. They also annotated response boundaries; agreement measures whether extracted start/end times fall within the annotation tolerance. The full protocol is in \Cref{app:human_study}.

As shown in \Cref{tab:human_validation}, both branches score above 4.0 on all three axes, with slightly higher means for reconstructed queries. Start-boundary agreement is 93--95\%, and end-boundary agreement is 88--90\%; response endings are harder to localize when streamers continue into related remarks. These results support the plausibility and temporal alignment of the extracted pairs. Additional acoustic and camera-motion analyses are in \Cref{app:data_analyses}.

\section{Experiments}
\label{sec:experiments}

\subsection{Problem Settings}
\noindent\textbf{Problem Formulation}. We formulate interactive audio-visual response generation as a VT2VA task: given the livestream context $\mathcal{C}$ together with a query $\mathcal{Q}$, the model is expected to generate an appropriate audio-visual response $\mathcal{Y}$, modeling $p(\mathcal{Y}\mid\mathcal{C},\mathcal{Q})$. Directly tackling this joint probability is challenging, as the model must simultaneously understand the livestream context, interpret the query intent, generate a textual response, and render a temporally coherent audio-video response.

Alternatively, we consider a two-stage generation pipeline. In the first stage, a vision-language model (VLM) functions as a planner that takes visual context $\mathcal{C}$ and the query $\mathcal{Q}$ as input, and produces an interaction-aware textual description $\mathcal{T}$. This description includes the relevant visual semantics of the scene and the intended response content of the streamer. In the second stage, an audio-video joint generation model takes the generated textual description $\mathcal{T}$ and context $\mathcal{C}$ as the conditioning signal and produces the final multimodal response, modeling $\mathcal{Y}$ through a two-step sampling process:
\begin{align}
p(\mathcal{Y}\mid\mathcal{C},\mathcal{Q})\approx \underbrace{p(\mathcal{Y}\mid\mathcal{T},\mathcal{C})}_{\textcolor{myblue}{\text{AV Co-Generator}}} \cdot \underbrace{p(\mathcal{T}\mid\mathcal{C},\mathcal{Q})}_{\textcolor{mygreen}{\text{Interaction Planner}}}
\end{align}
This decomposition allows us to model the semantic causality in natural interaction and the low-level audio-visual consistency separately. \Cref{app:single_stage} also evaluates a single-stage diffusion model for this task.

\begin{figure*}[!t]
    \centering
    \includegraphics[width=\linewidth]{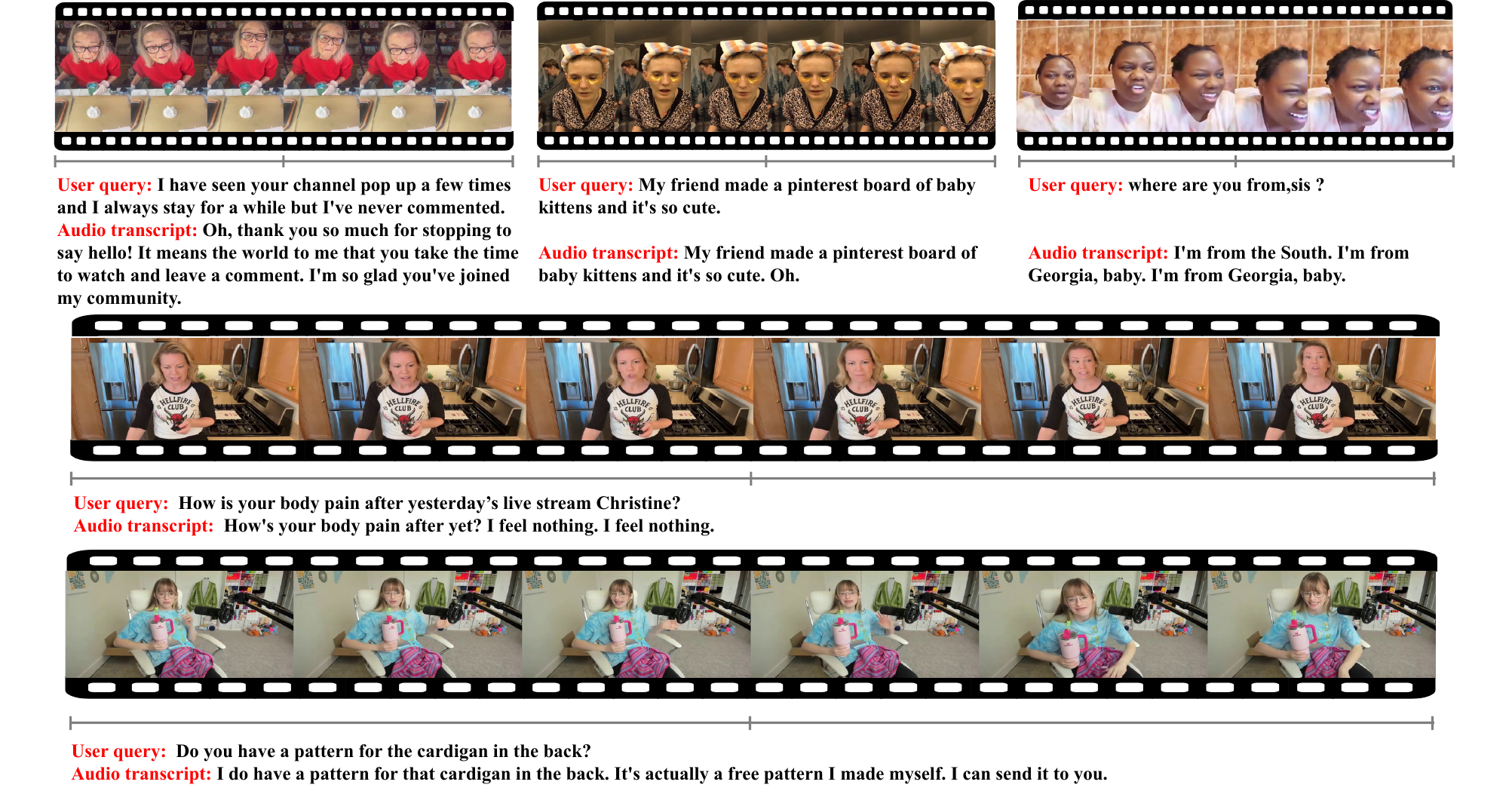}
    \caption{Qualitative responses from the fine-tuned Qwen3.5-9B planner and MOVA co-generator. Each example pairs a user query with generated video frames and the audio transcript, illustrating query-conditioned speech and visible actions.}
    \label{fig:qualitative_demos}
\end{figure*}

\subsection{Experimental Setup}
\label{sec:exp_setup}

\noindent\textbf{Training Setup.} We conduct decompositional experiments using our proposed data, isolating the learning objectives of the two-stage model and evaluating the effectiveness of our dataset on both stages separately. On one hand, we freeze the interaction planner and fine-tune the audio-visual co-generator $p(\mathcal{Y}\mid\mathcal{T},\mathcal{C})$ using $(\mathcal{C},\mathcal{T},\mathcal{Y})$ triples constructed from our dataset. This forces the generator to focus purely on visual consistency, audio quality, and temporal synchronization without being bottlenecked by flawed semantic inputs. On the other hand, we freeze the audio-visual co-generator and evaluate whether our dataset improves the upstream interaction planner. Fine-tuning the vision-language backbones on our data lets the planner learn to bridge the gap between the external user query $\mathcal{Q}$ and the interaction-aware textual description $\mathcal{T}$ necessary for coherent downstream generation. We use MOVA~\cite{yu2026mova} as the generator, with a first-frame context and eight-second training and generation clips (\Cref{app:training}).

\noindent\textbf{Evaluation Setup.} We reserve \textbf{100} examples from the real-query subset as held-out test cases. All 100 carry genuine live-chat queries rather than reconstructed ones, so the benchmark never measures a model against a synthetic trigger. The set is class-balanced across the four major domains and covers the range of query types, response durations, and human counts in the corpus. Crucially, the split is disjoint from training data at the \emph{channel and streamer} level, not merely at the clip level, which prevents a model from being rewarded for having memorized a particular streamer's appearance, voice, or verbal habits.

We introduce a three-dimensional evaluation protocol for interactive audio-visual responses, covering whether a response addresses the user, reacts appropriately, and conveys meaning through multiple modalities:
\begin{itemize}[nosep,leftmargin=10pt]
    \item \textbf{Relevance}: whether the response, in both spoken words and visuals, addresses what the trigger actually said, including the trigger's specific concrete details.
    \item \textbf{Appropriateness}: even an on-topic response may not be the right reaction. Appropriateness evaluates whether the generated audio-visual content matches the tone, emotion, and intent of the situation.
    \item \textbf{Expressiveness}: whether the response delivers its content in a rich multimodal way, with expressive gestures, facial expressions, and physical demonstrations, rather than as a flat read-aloud of text.
\end{itemize}
These dimensions are scored independently by Gemini-3.1-Pro~\cite{googledeepmind2026gemini31pro} and ten human raters on a 1--5 scale; OVRL is their mean. Gemini scores all 100 held-out examples, while human raters assess a sampled subset spanning the eight configurations in \Cref{tab:main}, blind to model identity, and additionally score video and audio quality. Values after $\pm$ are standard errors of the VLM means; \emph{pt.}/\emph{ft.} denote pretrained/fine-tuned models, and $\Delta$ rows measure stage-specific gains. For agreement analysis, we recompute Gemini scores on the same videos rated by humans (\Cref{app:human_contrasts}). \Cref{app:prompts,app:statistical_reporting} contain the evaluation prompts and statistical reporting details.

Besides these VLM- and human-based metrics, we also report conventional metrics for audio-visual joint generation, drawn from evaluation suites such as VBench~\cite{huang2024vbench}, AudioBox-Aesthetics~\cite{tjandra2025audiobox}, Verse-Bench~\cite{wang2025universe}, and other common metrics. See \Cref{app:evaluation,app:full_metrics} for details.

\begin{table*}[!t]
\centering
\small
\caption{\emph{Audio-visual quality with oracle captions.} (a) Zero-shot generator references, including LTX-2.3 with and without preceding audio-visual context (better values underlined). (b) MOVA before and after fine-tuning on InteracVid (better values in bold).}
\label{tab:av_eval}
\resizebox{\textwidth}{!}{%
\setlength{\tabcolsep}{4pt}
\renewcommand{\arraystretch}{1.05}
\begin{tabular}{lccccc?cccc?ccc}
\toprule
Method & \multicolumn{5}{c}{\cellcolor{red!30}{\textsc{Video quality}}} & \multicolumn{4}{c}{\cellcolor{yellow!30}{\textsc{Audio quality}}} & \multicolumn{3}{c}{\cellcolor{blue!30}{\textsc{A--V sync.}}} \\
\cmidrule(lr){2-6} \cmidrule(lr){7-10} \cmidrule(lr){11-13}
 & ID\,$\uparrow$ & SC\,$\uparrow$ & BC\,$\uparrow$ & MS\,$\downarrow$ & TF\,$\downarrow$ & CE\,$\uparrow$ & CU\,$\uparrow$ & PQ\,$\uparrow$ & CER\,$\downarrow$ & LC\,$\uparrow$ & LD\,$\downarrow$ & AVA\,$\uparrow$ \\
\midrule
\multicolumn{13}{l}{\textit{(a) Zero-shot generators: no task-specific fine-tuning}} \\
OVI~\cite{low2025ovi} & 0.905 & 0.971 & 0.969 & 0.008 & 0.011 & 5.112 & 5.576 & 6.002 & 0.251 & 1.984 & 8.483 & 0.240 \\
UniAVGen~\cite{zhang2025uniavgen} & 0.914 & 0.971 & 0.973 & 0.013 & 0.016 & 5.242 & 5.811 & 6.518 & 0.694 & 1.403 & 9.868 & 0.316 \\
UniVerse-1~\cite{wang2025universe} & 0.911 & 0.986 & 0.976 & 0.004 & 0.005 & 3.413 & 4.352 & 4.879 & 0.596 & 0.354 & 10.69 & 0.304 \\
\addlinespace[2pt]
\hdashline
\addlinespace[2pt]
LTX-2.3~\cite{lightricks2026ltx2} \emph{(first frame)} & \underline{0.888} & 0.969 & \underline{0.961} & \underline{0.011} & \underline{0.015} & 5.177 & 6.126 & 6.841 & 0.485 & \underline{2.066} & 8.308 & \underline{0.218} \\
LTX-2.3~\cite{lightricks2026ltx2} \emph{(+ preceding A/V)} & 0.882 & \underline{0.972} & 0.960 & \underline{0.011} & \underline{0.015} & \underline{5.665} & \underline{6.463} & \underline{7.126} & \underline{0.452} & 2.056 & \underline{7.466} & 0.181 \\
\midrule
\multicolumn{13}{l}{\textit{(b) Main comparison: same backbone, oracle caption}} \\
MOVA~\cite{yu2026mova} \emph{(pt.)} & 0.868 & 0.950 & 0.949 & 0.023 & 0.029 & 5.217 & 5.823 & 6.273 & 0.193 & 1.456 & 10.06 & 0.248 \\
\textbf{Ours} \emph{(ft.)} & \textbf{0.881} & \textbf{0.962} & \textbf{0.960} & \textbf{0.014} & \textbf{0.019} & \textbf{5.392} & \textbf{6.154} & \textbf{6.625} & \textbf{0.179} & \textbf{1.519} & \textbf{9.651} & \textbf{0.288} \\
\addlinespace[2pt]
\hspace{1em}$\Delta$\,\% & \textcolor{teal!75!black}{\scriptsize +1.5\%} & \textcolor{teal!75!black}{\scriptsize +1.3\%} & \textcolor{teal!75!black}{\scriptsize +1.2\%} & \textcolor{teal!75!black}{\scriptsize +39.2\%} & \textcolor{teal!75!black}{\scriptsize +34.7\%} & \textcolor{teal!75!black}{\scriptsize +3.4\%} & \textcolor{teal!75!black}{\scriptsize +5.7\%} & \textcolor{teal!75!black}{\scriptsize +5.6\%} & \textcolor{teal!75!black}{\scriptsize +7.1\%} & \textcolor{teal!75!black}{\scriptsize +4.3\%} & \textcolor{teal!75!black}{\scriptsize +4.1\%} & \textcolor{teal!75!black}{\scriptsize +16.3\%} \\
\bottomrule
\end{tabular}

}
\end{table*}

\subsection{Experimental Results}
\label{sec:exp_results}

We evaluate the semantic quality of two-stage interactive response generation in \Cref{tab:main}, combining VLM-judge and human scores. \Cref{tab:full_decomposition} reports the full VLM decomposition, including Qwen3-VL.

\noindent\textbf{Learning the AV Co-Generator with InteracVid.} We first use an off-the-shelf VLM as the interaction planner and fine-tune MOVA. As shown in \Cref{tab:main} and \Cref{tab:full_decomposition}, InteracVid is highly effective for teaching audio-visual co-generators: regardless of the VLM planner, the generator improves significantly and consistently on all metrics after fine-tuning, with $+1.41$ OVRL using Qwen3.5-9B, $+1.62$ using InternVL3.5-8B, and $+0.84$ using Qwen3-VL-8B as planner. The corresponding human-rated OVRL gains for Qwen3.5 and InternVL3.5 are $+1.04$ and $+1.68$, and human video-quality scores increase by $+1.60$ and $+2.49$.

\noindent\textbf{Learning the Interaction Planner with InteracVid.} Alternatively, we fix the fine-tuned MOVA co-generator and instruction-tune the off-the-shelf VLM-based planner. This leads to consistent semantic improvement across all three backbones, ranging from $+0.17$ to $+0.51$ VLM-rated OVRL (\Cref{tab:main,tab:full_decomposition}). Qwen3.5 and InternVL3.5 gain $+0.44$ and $+0.17$ under the VLM judge, and $+0.45$ and $+0.29$ under human scoring. The gains from tuning the planner are smaller than those from tuning the generator. We speculate that the planner's improvements are partly diluted during downstream generation: it affects the output through an intermediate textual description, and the generator's limitations can constrain how well that description is realized.

We also use oracle captions of the target response as a reference for generation with an accurate semantic plan. Fine-tuning the generator with these captions fixed yields only a marginal gain ($+0.09$ VLM-rated OVRL and $+0.14$ human-rated OVRL), since the caption already removes much of the ambiguity about the intended response. The remaining improvement mainly comes from adapting the generator to the livestream domain, including motion style, speech characteristics, and audio-visual synchronization. Oracle captions still outperform the best learned planner by $0.71$ VLM-rated OVRL and $0.38$ human-rated OVRL. This gap suggests that learning plausible response plans remains an important bottleneck within the current pipeline.

On the human-evaluation subset, judge and human configuration-level OVRL means correlate strongly (Pearson $r=0.964$ across eight configurations). \Cref{app:human_contrasts,app:counterfactual} report the full matched-video agreement analysis and counterfactual-query evaluation.

\noindent\textbf{Audio-Visual Generation Quality.} As a secondary evaluation, we assess whether InteracVid provides high-quality supervision to the audio-visual co-generator itself. Under oracle captions, fine-tuning MOVA improves all 12 metrics in \Cref{tab:av_eval}~(b), including CER ($0.193\rightarrow0.179$) and AV-Align ($0.248\rightarrow0.288$), demonstrating the value of the audio and video targets for improving generation fidelity and synchronization. We also report OVI~\cite{low2025ovi}, UniAVGen~\cite{zhang2025uniavgen}, UniVerse-1~\cite{wang2025universe}, and LTX-2.3~\cite{lightricks2026ltx2} as zero-shot reference baselines in block (a). The full two-stage evaluation with learned captions and the 26-metric analysis are in \Cref{app:two_stage_av,app:full_metrics}. Adding preceding A/V context to LTX-2.3 improves audio-quality scores and lip-sync distance, with mixed changes in other metrics; the detailed analysis is in \Cref{app:context_ablation}. \Cref{fig:qualitative_demos} shows qualitative results from the fine-tuned planner and generator, illustrating consistency, motion smoothness, and fidelity; further demonstrations and failure cases are in \Cref{app:demos}.

\section{Conclusion}
\label{sec:conclusion}

In this work, we introduce \textbf{InteracVid}, the first large-scale dataset for interactive audio-visual response generation collected from livestream videos. To construct the dataset at scale, we develop two complementary pipelines that leverage both natural live-chat interactions and reconstructed query--response pairs from reactive moments, and we validate with human raters that both branches yield causal, natural, and temporally complete interactions. Experiments on interactive audio-visual generation demonstrate that InteracVid provides effective supervision for both interaction planning and downstream audio-visual response generation, and a human study independently reproduces the ranking and conclusions obtained with our automatic judge. In particular, our results suggest that learning plausible interactive response planning remains a key bottleneck for future multimodal interactive agents.
We hope InteracVid will facilitate future research on multimodal assistants and interactive agents capable of grounded, expressive, and interaction-aware audio-visual responses.

{\small
\bibliographystyle{ieeenat_fullname}
\bibliography{main}}

\ifincludesupplement
  \beginsupplement
  \section*{LLM Usage Statement}
Generative AI tools were used for dataset curation, automated evaluation,
and auxiliary writing and coding tasks.
Seed-1.6~\citep{bytedance2025seed16} was used for transcript normalization,
query--response matching, discourse-role classification, query reconstruction,
multimodal quality filtering, and auxiliary audio-visual captioning in the
dataset curation pipeline described in Section~\ref{sec:curation}.
Seed-ASR~\citep{bai2024seedasr} was used for speech transcription and
transcript refinement, while Qwen2-Audio~\citep{chu2024qwen2audio} was used
for acoustic auditing and annotation.
Gemini-3.1-Pro~\citep{googledeepmind2026gemini31pro} was used for automated
evaluation of generated audio-visual responses, as described in
Section~\ref{sec:experiments}.
ChatGPT assisted with language editing and drafting portions of the
manuscript, while OpenAI Codex assisted in generating parts of the project
code. AI-assisted manuscript text and code were carefully reviewed,
verified, and revised by the authors.
Automated annotations and evaluations were assessed through the
quality-control procedures and sampled human evaluations reported in
the paper.
The authors formulated the research questions, designed the methodology
and experimental protocols, interpreted the results, and drew the
scientific conclusions. The authors take full responsibility for the
content and conclusions of this paper.
  \startcontents[appendix]
  \section*{Supplementary Contents}
\begingroup
\hypersetup{linkcolor=black}
\printcontents[appendix]{app}{1}[2]{}
\endgroup
\par\medskip

\section{Details on Dataset Curation Pipeline}
\label{app:curation}

\Needspace{6\baselineskip}
\subsection{Class-Balanced Crawling}
\label{app:crawling}

To construct a diverse livestream corpus, we adopt a class-balanced crawling strategy. Rather than collecting videos from a small number of high-frequency search queries, we define a taxonomy of livestream scenarios and crawl videos independently for each subclass. Our taxonomy is organized into four high-level groups.

The first group consists of conversation-centered streams, including casual chatting, Q\&A, interviews, podcasts, debates, and viewer-submission reviews. These videos are included because audience reactions are often tightly coupled with spoken content, opinions, jokes, personal stories, and direct streamer-viewer interaction.

The second group consists of lifestyle, object-centric, and sensory streams, including unboxing, shopping hauls, live sales, beauty routines, mukbang, ASMR, crafting, and creative construction. These videos contain visible objects, changing states, and evaluative moments. Viewer comments may refer to product appearance, manual progress, taste, texture, object comparison, or visual transformation.

The third group consists of procedural and embodied activity streams, including fitness, yoga, dance practice, cooking, baking, and food preparation. These videos contain temporally structured actions with clear intermediate states and outcomes. Viewer comments may refer to action correctness, physical difficulty, procedural progress, or the final result.

The fourth group consists of digital and screen-based streams, including live gameplay, coding and software tutorials, and digital art or creative editing. These videos contain screen-space actions and user-interface state, and viewer comments often refer to on-screen events, tool choices, or the creator's next move.

For each subclass, we construct search queries by combining content-specific keywords with livestream-oriented terms such as \textit{live}, \textit{livestream}, \textit{stream replay}, and \textit{VOD}. We retain at most 20 videos per query and apply basic filters on duration, resolution, language, audio availability, and subtitle availability. The final crawling taxonomy and corresponding keywords are shown in \Cref{tab:crawling_keywords_appendix}.

\begin{table}[!tp]
\centering
\small
\setlength{\tabcolsep}{3.5pt}
\renewcommand{\arraystretch}{1.15}

\caption{Taxonomy, representative keywords, and quotas for class-balanced crawling across four livestream domains.}
\label{tab:crawling_keywords_appendix}
\begin{tabular}{@{}
    >{\raggedright\arraybackslash}p{0.15\linewidth}
    >{\raggedright\arraybackslash}p{0.20\linewidth}
    >{\raggedright\arraybackslash}p{0.52\linewidth}
    >{\centering\arraybackslash}p{0.07\linewidth}
@{}}

\toprule
\textbf{Group} & \textbf{Scenario} & \textbf{Representative keywords} & \textbf{Quota} \\
\midrule

\rowcolor{gray!12}
\multicolumn{4}{@{}l}{\textbf{Conversation-centered streams}} \\

Conversation
& Casual Chat and Q\&A
& Chatting, live chat, Q\&A, question and answer, ask me anything, storytime, get ready with me, GRWM, GRWM for school, late night chat, real talk, mental health check-in, wakeup chat, spoiler, ASMR chat
& 15\% \\

Conversation
& Debate, Interview, and Viewer Review
& Online debate, interview, podcast, Discord debate, couple Q\&A, sibling live, watching fan videos, reviewing viewer submissions, roasting viewers
& 5\% \\

\rowcolor{gray!12}
\multicolumn{4}{@{}l}{\textbf{Lifestyle and object-centric streams}} \\

Lifestyle / Object
& Unboxing and Mystery Boxes
& Unboxing, box opening, unboxing haul, haul unboxing, sneaker unboxing, bag unboxing, Amazon mystery box, eBay mystery box, subscription box opening, blind box opening
& 8\% \\

Lifestyle / Object
& Shopping Hauls and Live Sales
& Clothing haul, try-on haul, Zara haul, SHEIN haul, thrift haul, crystal claim sale, crystal sale, clothing claim sale, live auction, claim sale, jewelry sale, bin sale, jewelry auction, handbag sale, closet cleanout, makeup haul
& 8\% \\

Lifestyle / Object
& Beauty, Skincare, and Hair Styling
& Makeup, skincare routine, full face makeup, eye makeup, eyeliner, makeup tutorial, lipstick swatching, morning routine, night routine, doing nails, acrylic nails tutorial, gel nails, hair styling, hair cutting, cut my hair, dye my hair, braiding hair
& 8\% \\

Lifestyle / Object
& Mukbang and Beverage Tasting
& Mukbang, eating show, pizza mukbang, fried chicken mukbang, noodle mukbang, spicy food challenge, dessert mukbang, tteokbokki mukbang, burger mukbang, wine tasting, coffee tasting, beer drinking
& 8\% \\

Lifestyle / Object
& Crafting, Building, and DIY
& Lego build, Lego Star Wars build, Gundam build, Gunpla build, keyboard build, making something, live soldering, crochet with me, knitting, embroidery, clay sculpting, pottery throwing, painting miniatures, watercolor, journaling, scrapbooking, origami, jewelry making, resin art, candle making
& 8\% \\

Lifestyle / Object
& ASMR Interaction
& ASMR chat, ASMR eating, ASMR tapping, reading chat, ASMR ear cleaning
& 3\% \\

\rowcolor{gray!12}
\multicolumn{4}{@{}l}{\textbf{Procedural and embodied streams}} \\

Procedural / Embodied
& Fitness, Yoga, and Dance
& Workout, workout tutorial, full body workout, yoga flow, morning yoga, Pilates, stretching routine, Zumba class, dance practice, abs workout, pushup challenge, pushup, squat, squat challenge, plank, plank challenge, home gym workout, treadmill, treadmill walking
& 10\% \\

Procedural / Embodied
& Cooking and Baking
& Cooking, baking, cook with me, decorating cake, making pasta, making pizza, making sushi, meal prep, vegan cooking, BBQ, making coffee
& 10\% \\

\rowcolor{gray!12}
\multicolumn{4}{@{}l}{\textbf{Digital and screen scenario streams}} \\

Digital / Screen
& Gaming and Live Gameplay
& gaming, gameplay, let's play, speedrun, Minecraft, Fortnite, Valorant, League of Legends, walkthrough, game challenge
& 6\% \\

Digital / Screen
& Software, Coding, and Screen Tutorials
& Coding, programming, code with me, software tutorial, web design, Photoshop, Blender tutorial, screen share
& 5\% \\

Digital / Screen
& Digital Art and Creative Editing
& Digital art livestream, drawing, Procreate, animation, 3D modeling, Blender sculpting live, video editing live, graphic design, concept art, live illustration, thumbnail design
& 6\% \\

\bottomrule
\end{tabular}

\end{table}

\Needspace{6\baselineskip}
\subsection{Sentence-Level Semantic Timeline}
\label{app:timeline}

Raw subtitle tracks are often fragmented into short clauses and may not correspond to complete semantic units. We therefore normalize subtitles into a sentence-level timeline. Given a raw subtitle sequence, we first split it into chunks, with each chunk 100 lines of subtitles. The choice of this number is ablated to balance processing speed and effect. Each chunk is processed by an instruction-tuned language model that merges fragments, restores punctuation, removes repeated subtitle artifacts, and preserves approximate timestamps.

The output is a sequence
\begin{equation}
    S = \{(s_i,t_i)\}_{i=1}^{N},
\end{equation}
where $s_i$ is a complete sentence or short utterance and $t_i$ is the estimated timestamp of the start and end. This representation is used by both the real-data and synthetic-data branches. It allows the pipeline to judge whether a sentence is a proactive statement, a direct answer, a continuation of a previous answer, or a topic shift.

\Needspace{6\baselineskip}
\subsection{Real-Query Branch}
\label{app:real_branch}

For videos with time-stamped live-chat or danmaku metadata, we construct candidate query-response pairs by matching user comments with subsequent streamer utterances. For each transcript sentence $(s_i,t_i)$, we collect comments in a preceding window:
\begin{equation}
    D_i = \{d_j \mid t_i - t_{\mathrm{window}} \leq \tau_j \leq t_i\},
\end{equation}
where $d_j$ is a comment and $\tau_j$ is its timestamp. We set $t_{\mathrm{window}}=60s$ in the final pipeline.

Temporal proximity alone is insufficient because many comments are ignored, repeated, or unrelated to the streamer response. We therefore apply semantic correspondence filtering. Given a candidate comment, the local transcript context, and optionally sampled video frames, a vision-language model determines whether the comment plausibly triggers the response. We retain a pair only when the model predicts a direct or contextually valid relation, such as answering a question, reacting to a compliment, following a request, clarifying a visible object, or responding to feedback.

After identifying the trigger sentence, we extend the response span by inspecting the following 5 transcript sentences. Consecutive sentences are retained if they continue the same response, and the span is terminated when the streamer changes topic, returns to monologue, or begins responding to another comment.

\Needspace{6\baselineskip}
\subsection{Reconstructed-Query Branch}
\label{app:synthetic_branch}

Many livestream replays and VOD videos do not preserve live-chat metadata on YouTube. To make use of these videos, we construct the missing query condition in the reverse direction. We first identify response-like spans from the sentence-level timeline. Each sentence or short span is classified into one of the following discourse roles:
\begin{itemize}
    \item \textbf{Proactive monologue}: the streamer introduces or continues a topic without apparent external stimulus.
    \item \textbf{Reactive response}: the streamer appears to answer, clarify, demonstrate, or react to an implicit user-side stimulus.
    \item \textbf{Continuation}: the sentence continues a response that began earlier.
\end{itemize}

For spans classified as reactive responses, we ask a vision-language model to observe the local video context and response transcript, then generate a plausible user query that could have triggered the response. The generated query must be concise, natural, and grounded in the visible or spoken context.

\Needspace{6\baselineskip}
\subsection{Response Clip Extraction}
\label{app:clip_extraction}

Once a response span is determined, we extract the corresponding audio-video clip from the original video. Due to the native inaccuracy of the timestamps in video subtitles, we expand the cutting window to ensure that the full response is contained in the resulting clip. We add a 0.5-second offset before the start and after the end, to preserve natural speech boundaries. Clips shorter than 3 seconds are removed from the training data. \Cref{sec:human_validation} reports how often the resulting boundaries agree with human annotation.

The preceding context $\mathcal{C}$ is extracted from the interval immediately before the response. Depending on the downstream model, $\mathcal{C}$ can be represented as a short video clip, a set of frames, the first frame, the previous transcript sentences, or a combination of these signals. In the main experiments, we use the first frame image as context for simplicity and compatibility with current audio-video generation models.

\Needspace{6\baselineskip}
\subsection{Multimodal Quality Filtering}
\label{app:quality_filtering}

We apply multimodal filtering to remove samples that are unsuitable for audio-visual response generation. The filters include:

\noindent\textbf{Transcript-visual alignment.}
A vision-language model checks whether the response transcript is grounded in the visible scene. Samples are removed when the speech is unrelated to the video, when the video only shows a static placeholder, or when the response relies entirely on unavailable off-screen context.

\noindent\textbf{Scene continuity.}
We want the audio-visual response to be largely consistent and natural, without abrupt scene transitions. We therefore detect abrupt scene transitions within the response clip using PySceneDetect. Clips with large discontinuities are removed because they break the temporal coherence between context and response.

\noindent\textbf{Language.}
The same vision-language pass flags whether the primary spoken language of the response is English, and clips that are not primarily English are discarded. Together with the English crawling keywords, this is why more than 99\% of the released corpus is English; the consequences are discussed in \Cref{sec:statistics} and \Cref{app:ethics_release}.

\noindent\textbf{Updated ASR.}
As noted above, subtitle timestamps and content are natively inaccurate. Errors in the subtitle-derived text severely harm audio-visual generation quality during training. We therefore rerun ASR with Seed-ASR~\cite{bai2024seedasr} on the extracted response clip and compare the result with the original subtitle-derived transcript. Samples are removed when the two transcripts differ by more than a 40\% word error rate. The updated ASR output contains precise timestamps for every word, which lets us crop the clips to an arbitrary generation length.

\noindent\textbf{What these filters do not check.}
Every filter above is anchored on speech: the relevance check compares the spoken transcript against the visuals, and the ASR check validates that transcript. Nothing verifies whether co-occurring non-speech audio --- background music, sound effects, or ambient noise --- is semantically related to the scene, because the vision-language model used for filtering is optimized for visual and textual reasoning rather than fine-grained acoustic understanding. We quantify the resulting exposure in \Cref{app:audio_audit} and release per-clip acoustic annotations so that downstream users can filter on this axis explicitly.

\Needspace{6\baselineskip}
\subsection{Auxiliary Caption Generation}
\label{app:captioning}

For each retained response clip, we generate an auxiliary audio-visual caption $\mathcal{T}$. The caption is designed to describe both the speech content and the visible behavior. It typically contains: (1) a concise description of the scene; (2) the streamer action or reaction; (3) the spoken response or its semantic summary; and (4) any salient object or screen interaction. This caption is used as an intermediate condition in the two-stage generation experiments. The caption is generated given both the video clip and the updated ASR results, making sure that it contains the correct speech information.

\Needspace{6\baselineskip}
\subsection{LLM / VLM Usage}
\label{app:llm_usage}

For large-scale dataset construction, we use Seed-1.6~\cite{bytedance2025seed16} as the LLM and VLM, because its API cost and achievable concurrency make corpus-scale filtering feasible. ASR is performed with Seed-ASR~\cite{bai2024seedasr}. The acoustic audit in \Cref{app:audio_audit} uses Qwen2-Audio~\cite{chu2024qwen2audio}, which unlike the curation VLM is trained for fine-grained acoustic understanding. For evaluation, we use Gemini-3.1-Pro~\cite{googledeepmind2026gemini31pro} for its strong native audio-visual understanding. All model versions are fixed across the experiments reported here, and all prompts are listed in \Cref{app:prompts}.

\Needspace{6\baselineskip}
\subsection{Dataset Format}
\label{app:dataset_format}

Each sample is stored with the fields listed in \Cref{tab:dataset_format}.

\begin{table}[t]
\centering
\small
\renewcommand{\arraystretch}{1.1}
\caption{Released dataset fields. Each record contains source identifiers and derived annotations rather than redistributed media; see \Cref{app:ethics_release}.}
\label{tab:dataset_format}
\begin{adjustbox}{max width=\linewidth}
\begin{tabular}{ll}
\toprule
Field & Description \\
\midrule
\texttt{sample\_id} & Unique sample identifier \\
\texttt{video\_id} & Source video identifier \\
\texttt{channel\_id} & Source channel identifier, used to enforce split disjointness \\
\texttt{search\_key} & Crawling keywords for scenario category \\
\texttt{query} & Real or reconstructed user query \\
\texttt{query\_source} & \texttt{real} or \texttt{reconstructed} \\
\texttt{context\_start}, \texttt{context\_end} & Time range for preceding context \\
\texttt{response\_start}, \texttt{response\_end} & Time range for response clip \\
\texttt{response\_transcript} & Refined ASR transcript with word-level timestamps \\
\texttt{av\_caption} & Auxiliary audio-visual caption \\
\texttt{audio\_pattern} & Acoustic composition labels from the audit in \Cref{app:audio_audit} \\
\texttt{camera\_motion} & \texttt{stationary}, \texttt{moving}, or \texttt{navigation} (\Cref{app:camera_motion}) \\
\texttt{split} & \texttt{train}, \texttt{val}, or \texttt{test} \\
\texttt{metadata} & Resolution, duration, language, filters, and other attributes \\
\bottomrule
\end{tabular}
\end{adjustbox}
\end{table}

\section{Additional Dataset Analyses}
\label{app:data_analyses}

\Needspace{6\baselineskip}
\subsection{Human Validation Protocol}
\label{app:human_study}

\Cref{sec:human_validation} reports a human study on the quality of the extracted interactions. This supplementary section documents the protocol.

\noindent\textbf{Sampling.} Query--response pairs were sampled uniformly at random within a stratification over (branch $\times$ domain), where branch is \texttt{real} or \texttt{reconstructed} and domain is one of the four major scenario groups. Stratification ensures that neither branch is evaluated on an easier domain mix. We observed no significant difference in ratings across domains, so \Cref{tab:human_validation} aggregates over them.

\noindent\textbf{Raters and load.} Ten raters each evaluated ten pairs, giving 100 judgments per branch. Raters were shown the query, the response clip with audio, and the surrounding context; they were not told which branch a query came from, so the comparison between real and reconstructed queries is blind.

\noindent\textbf{Rating axes.} Each pair was scored from 1 to 5 on:
\begin{itemize}[nosep,leftmargin=12pt]
    \item \textbf{Causality}: could this query plausibly have caused this response? A score of 1 means the two are unrelated; 5 means the response is clearly an answer or reaction to the query.
    \item \textbf{Naturalness}: does the query read like a message a real viewer would type into live chat? A score of 1 means it reads as machine-generated or implausibly formal.
    \item \textbf{Completeness}: does the clip contain the whole response, without truncating it or trailing into unrelated speech?
\end{itemize}

\noindent\textbf{Span validation.} In a second pass, raters saw the preceding and succeeding context and marked where they believed the response actually began and ended. We then compared these annotations against our automatically extracted boundaries and counted a boundary as correct when it fell within the predefined tolerance. Reported agreement is per-boundary, which is why start and end accuracies differ.

\Needspace{6\baselineskip}
\subsection{Acoustic Composition Audit}
\label{app:audio_audit}

Because the curation filters are speech-anchored (\Cref{app:quality_filtering}), we audit the audio track with Qwen2-Audio~\cite{chu2024qwen2audio}, which labels each clip with the set of sound categories it contains. \Cref{tab:audio_audit} reports the result.

\begin{table}[t]
\centering
\small
\caption{Qwen2-Audio audit. Left: coarse acoustic composition and branch-specific speech-only rates. Right: non-speech sound prevalence across all clips; categories may overlap.}
\label{tab:audio_audit}
\begin{subtable}[t]{0.55\linewidth}
\centering
\caption{Coarse composition.}
\setlength{\tabcolsep}{5pt}
\renewcommand{\arraystretch}{1.1}
\begin{adjustbox}{max width=\linewidth}
\begin{tabular}{lccc}
\toprule
Audio pattern & All & Real & Recon. \\
\midrule
Speech only            & 57.6\% & 62.8\% & 52.4\% \\
Speech + other sounds  & 42.0\% & -- & -- \\
Non-speech only        & \phantom{0}0.32\% & -- & -- \\
\midrule
\emph{Any speech}      & \textbf{99.68\%} & -- & -- \\
\bottomrule
\end{tabular}
\end{adjustbox}
\end{subtable}
\hfill
\begin{subtable}[t]{0.42\linewidth}
\centering
\caption{Most frequent non-speech sounds.}
\setlength{\tabcolsep}{5pt}
\renewcommand{\arraystretch}{1.1}
\begin{adjustbox}{max width=\linewidth}
\begin{tabular}{lc}
\toprule
Sound category & Share of clips \\
\midrule
Music                     & 27.6\% \\
Mechanical sounds         & 10.7\% \\
Human non-speech sounds   & \phantom{0}9.90\% \\
Electronic sound effects  & \phantom{0}3.36\% \\
Animal sounds             & \phantom{0}2.82\% \\
\bottomrule
\end{tabular}
\end{adjustbox}
\end{subtable}
\end{table}

Two facts matter for how the dataset should be used. First, speech is present in $99.68\%$ of clips and is therefore the dominant semantic channel of the response, which confirms the speech-mediated interaction setting InteracVid targets: a response's meaning is essentially always carried by speech rather than by unrelated background audio. Second, $42\%$ of clips carry additional sound, most often music. The released \texttt{audio\_pattern} field lets users exclude such clips, stratify evaluation by acoustic composition, or move the acoustic attributes into the conditioning prompt so that they are explained rather than treated as noise; modelling the semantic relationship between non-speech audio and the interaction is a natural extension of these annotations.

\Needspace{6\baselineskip}
\subsection{Camera Motion Estimation}
\label{app:camera_motion}

Livestream footage is dominated by fixed setups: face-to-camera framing, tabletop views, and screen capture are the most common formats, and they are also the primary application scenarios for interactive audio-visual generation. To characterize how much of the corpus goes beyond fixed-camera footage, we estimate camera motion for each response clip by combining a structure-from-motion backbone~\cite{wang2025vggt} with sparse Kanade--Lucas--Tomasi (KLT) feature tracking. The two signals are complementary: the former yields metric-scale camera translation and rotation, while the latter distinguishes a moving camera from a static camera observing a moving subject.

We assign each clip one of three labels:
\begin{itemize}[nosep,leftmargin=12pt]
    \item \textbf{Stationary}: feature tracks are stationary in at least $75\%$ of frames.
    \item \textbf{Navigation-like}: the estimated camera translation diameter is at least $0.15$, or the estimated rotation is at least $5^{\circ}$, or dynamic background tracks are present in at least $50\%$ of frames.
    \item \textbf{Moving}: everything else, that is, intermediate camera movement such as handheld drift, reframing, or panning.
\end{itemize}

\begin{table}[t]
\centering
\small
\setlength{\tabcolsep}{8pt}
\renewcommand{\arraystretch}{1.1}
\caption{Camera dynamics of response clips, as a percentage of each branch. Labels are assigned automatically from structure-from-motion and sparse feature tracking.}
\label{tab:camera_motion}
\begin{adjustbox}{max width=\linewidth}
\begin{tabular}{l ccc}
\toprule
Query branch & Stationary & Moving & Navigation-like \\
\midrule
Real          & 82.87 & 12.85 & 4.28 \\
Reconstructed & 86.84 & \phantom{0}8.80 & 4.36 \\
\bottomrule
\end{tabular}
\end{adjustbox}
\end{table}

\Cref{tab:camera_motion} reports the resulting distribution. Roughly $84$--$87\%$ of clips are stationary, which reflects the reality of livestream production. A further $9$--$13\%$ exhibit intermediate camera movement, and the navigation-like share is consistent across branches at $4.3\%$, corresponding to more than $19$K clips that cover walking tours, kitchen and workshop navigation, and handheld outdoor streams. We label rather than filter these clips so that they can be selected or excluded deliberately.

\section{Details on Training}
\label{app:training}

\Needspace{6\baselineskip}
\subsection{Audio-Visual Co-Generator Fine-Tuning}
\label{app:training_avgenerator}

We fine-tune the audio-visual co-generator using triples $(\mathcal{C},\mathcal{T},\mathcal{Y})$, where $\mathcal{C}$ is the visual context, $\mathcal{T}$ is the auxiliary audio-visual caption, and $\mathcal{Y}$ is the response clip. For simplicity and compatibility with current audio-video generation models, in the main experiments, $\mathcal{C}$ is represented by the first frame, following common practice. The model is initialized from MOVA~\cite{yu2026mova} and fine-tuned using LoRA with rank 16. As MOVA supports only 8-second videos, we crop all training videos to 8 seconds. We train with AdamW~\cite{loshchilov2019adamw}, a constant learning rate of 1e-4, and a local batch size of 1 for 1600 steps on 32 H20 GPUs.

We retain MOVA's flow-matching objective~\cite{yu2026mova}. For modality $m\in\{v,a\}$, let $z_0^m$ be the video or audio latent of the response and $z_t^m=(1-t)z_0^m+t\epsilon^m$ its noisy interpolation at noise level $t\in[0,1]$, where $\epsilon^m\sim\mathcal{N}(0,I)$. The objective regresses the target velocity $\epsilon^m-z_0^m$:
\begin{equation}
    \mathcal{L}_{\mathrm{gen}}
    =
    \mathbb{E}_{\mathcal{Y},\epsilon,t}
    \left[
    \sum_{m\in\{v,a\}}\lambda_m
    \left\|
    v_{\theta}^{m}(z_t^v,z_t^a,t,\mathcal{C},\mathcal{T})
    -(\epsilon^m-z_0^m)
    \right\|_2^2
    \right],
\end{equation}
where $v_{\theta}^{m}$ is the predicted velocity for modality $m$, and $\lambda_v$ and $\lambda_a$ weight the video and audio losses, respectively. The first-frame context $\mathcal{C}$ and auxiliary caption $\mathcal{T}$ provide the conditioning signals.

\Needspace{6\baselineskip}
\subsection{Interaction Planner Fine-Tuning}
\label{app:training_planner}

We fine-tune vision-language models to predict the interaction-aware textual description $\mathcal{T}$ from the context $\mathcal{C}$ and query $\mathcal{Q}$. The planner is trained on pairs:
\begin{equation}
    (\mathcal{C},\mathcal{Q}) \rightarrow \mathcal{T}.
\end{equation}
This stage is not intended to evaluate text-only dialogue generation in isolation. Instead, it tests whether the dataset can teach a model to transform a user query and visual context into a useful conditioning signal for downstream audio-video generation.

We train three models: Qwen3-VL-8B~\cite{bai2025qwen3vl}, InternVL-3.5-8B~\cite{wang2025internvl35}, and Qwen3.5-9B~\cite{yang2025qwen3} using LoRA fine-tuning with rank 16, learning rate 2e-4, cosine scheduling, global batch size 16, and gradient accumulation step 4 for 2 epochs. Evaluation is done on the 100 held-out examples. The target description is the auxiliary audio-visual caption generated during curation. We add a system prompt before the context and query input, to improve model performance:
\begin{lstlisting}[style=promptstyle]
You are an expert video captioner specialized in describing live stream content. Given a first frame image from a video and a live chat comment, generate a detailed description of what happens in the video, focusing on how the streamer responds to the comment.

Your caption should be a detailed multi-sentence paragraph (typically 3-8 sentences) that includes:

1. **Scene Description**: The physical environment, setting, lighting, background elements, and any visible objects or UI elements (game screens, chat overlays, etc.)

2. **People**: Describe the appearance, clothing, position, and distinctive features of anyone visible in the frame

3. **Actions & Gestures**: Movement, body language, hand gestures, and physical interactions with objects or other people

4. **Speech**: What the person says, enclosed in square brackets [like this]. Only include actual spoken dialogue that can be inferred from the context.

5. **Reactions**: How the streamer responds to the chat comment - their emotional tone, engagement level, and any specific actions taken in response

Write in present tense. Be specific and descriptive. Describe what you can see in the frame and what actions/movements occur. Place any spoken dialogue in square brackets within the natural flow of the description.
\end{lstlisting}
During inference, the generated description is passed to the fixed audio-visual co-generator.

\section{Single-Stage Generation Attempt}
\label{app:single_stage}

We also explored a single-stage formulation that directly maps the interaction condition to the final audio-visual response via finetuning a single diffusion model, asking it to model:
\begin{equation}
    p(\mathcal{Y} \mid \mathcal{C}, \mathcal{Q}),
\end{equation}
where $\mathcal{C}$ is the preceding context hereby simplified as the first frame image, $\mathcal{Q}$ is the user query, and $\mathcal{Y}$ is the response video and audio. This formulation is conceptually attractive because it avoids an intermediate textual planning stage and allows the model to jointly learn semantic response selection and low-level audio-visual realization. It also examines the question of whether diffusion models can possess audio-visual semantic reasoning abilities.

\noindent\textbf{Experimental setup.}
We instantiate the single-stage baseline using the same configuration as in the AudioVisual Co-Generator in our two-stage experiment. We use the pretrained MOVA as the base generation model. The model receives the first frame and a modified prompt based on the user query as input:
\begin{lstlisting}[style=promptstyle]
    Generate a video of the streamer answering user query: {user_query}
\end{lstlisting}
and is also trained to generate 8-second audio-visual responses. We train with AdamW~\cite{loshchilov2019adamw}, constant learning rate 1e-4, with local batch size of 1, and LoRA~\cite{hu2022lora} rank 16 for 800 steps on 16 H20 GPUs.

\begin{table}[t]
  \centering
  \caption{Single-stage response generation evaluated by Gemini-3 Pro. OVRL averages relevance, appropriateness, and expressiveness; best values are bold.}
  \label{tab:single-stage}
  \setlength{\tabcolsep}{4pt}
  \renewcommand{\arraystretch}{1.10}
  \small
  \begin{adjustbox}{max width=\linewidth}
\begin{tabular}{l l cccc}
    \toprule
    Setting & Model & Relev. & Approp. & Expr. & Ovrl \\
    \midrule
    Oracle + stage 2 & MOVA~\cite{yu2026mova} \emph{(ft.)}
      & \textbf{4.97\,$\pm$\,0.15} & \textbf{4.54\,$\pm$\,0.083} & \textbf{3.41\,$\pm$\,0.09} & \textbf{4.31} \\
    Two-stage model & + Qwen3.5-9B~\cite{yang2025qwen3} \emph{(ft.)}
      & 3.91\,$\pm$\,0.16    & 3.63\,$\pm$\,0.19      & 3.25\,$\pm$\,0.09      & 3.60    \\
    Single-stage model
      & MOVA~\cite{yu2026mova} \emph{(ft.\ on trigger)} & 1.57\,$\pm$\,0.09 & 1.39\,$\pm$\,0.05 & 1.50\,$\pm$\,0.07 & 1.49 \\
    \bottomrule
  \end{tabular}
\end{adjustbox}
\end{table}

\begin{figure}[t]
    \centering
    \includegraphics[width=\linewidth]{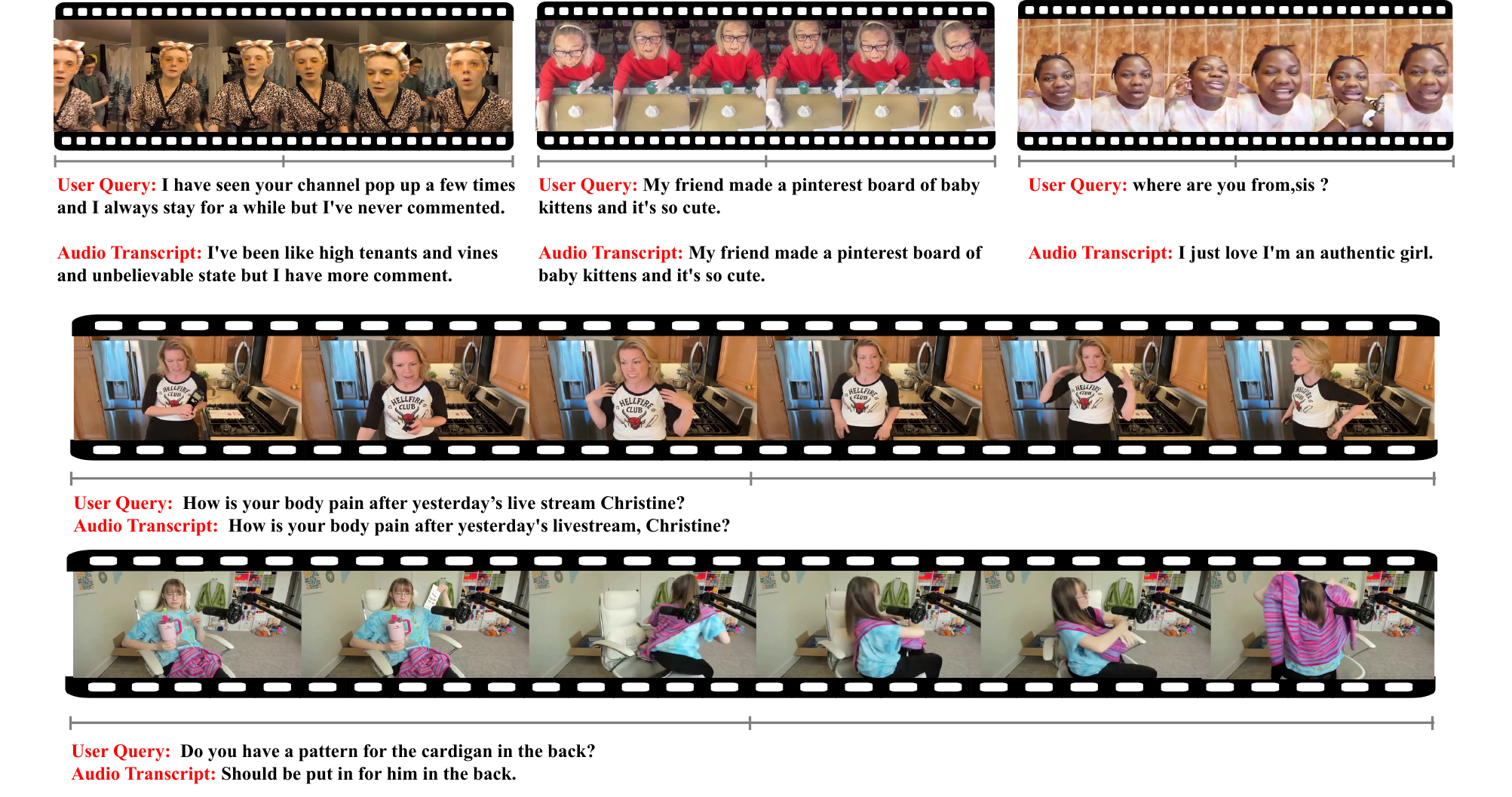}
    \caption{Single-stage diffusion model output. We train the MOVA model to generate the responding audio-visual interaction directly conditioned on the context and query.}
    \label{fig:single-stage}
\end{figure}
\noindent\textbf{Observed behavior.}
The single-stage model learns low-level appearance and motion priors, but does not maintain the interaction relation between the query and the response. The generated speech is generally intelligible, but falls back on shallow patterns such as repeating the question (which does happen in real livestreams) or lifting individual words from the query into the response. We evaluated the results using the same VLM-based semantic metrics, as shown in \Cref{tab:single-stage}. Demonstrations are shown in \Cref{fig:single-stage}.

\noindent\textbf{Discussion.}
These results show that current diffusion backbones are not strong enough for direct end-to-end learning from $\mathcal{C}, \mathcal{Q}$ to $\mathcal{Y}$, which is why we adopt the two-stage setting in the main experiments. The first stage predicts an interaction-aware textual description, while the second stage synthesizes the audio-visual response. This decomposition isolates high-level response planning from low-level audio-video rendering and provides a more diagnostic evaluation of how our dataset benefits each component.

\section{Details on Evaluation}
\label{app:evaluation}

\Needspace{12\baselineskip}
\subsection{Full VLM-Judge Decomposition}
\label{app:full_decomposition}

The main paper combines the automatic VLM-judge results with the human evaluation and therefore follows the row order of the human study. This table gives the complete automatic-judge decomposition, including the Qwen3-VL planner that has no corresponding human-evaluation rows.

\begin{table}[!h]
\centering
\small
\setlength{\tabcolsep}{5pt}
\renewcommand{\arraystretch}{1.05}
\caption{\emph{Full semantic evaluation by the automatic VLM judge}. This is the complete version of the VLM block in \Cref{tab:main}, including Qwen3-VL. Scores are 1--5 ratings; values after $\pm$ are standard errors of the mean.}
\label{tab:full_decomposition}
\begin{adjustbox}{max width=\linewidth}
\begin{tabular}{l c c c c c c}
\toprule
Model & \cellcolor{myblue!30}{Plan.\ ft.} & \cellcolor{mygreen!30}{Gen.\ ft.} & \cellcolor{green!30}{Relev.}\,$\uparrow$ & \cellcolor{green!30}{Approp.}\,$\uparrow$ & \cellcolor{green!30}{Expr.}\,$\uparrow$ & \cellcolor{green!30}{OVRL}\,$\uparrow$ \\
\midrule
\multicolumn{7}{l}{\textit{(a) Benefits of fine-tuning the AV Co-Generator}} \\
\multirow{3}{*}{Oracle (GT)}
              & -- & \ding{55}                                   & 4.89\,$\pm$\,0.042           & \textbf{4.61\,$\pm$\,0.108}  & 3.15\,$\pm$\,0.10            & 4.22 \\
              & -- & \textbf{\ding{51}}                          & \textbf{4.96\,$\pm$\,0.03}   & 4.54\,$\pm$\,0.083           & \textbf{3.41\,$\pm$\,0.09}   & \textbf{4.31} \\
              & \multicolumn{2}{c}{$\Delta$\;\textcolor{black!50}{\scriptsize(+Gen. ft.)}}                  & \textcolor{teal}{+0.07}      & \textcolor{black!45}{$-$0.07} & \textcolor{teal}{+0.26}      & \textcolor{teal}{+0.09}      \\
\addlinespace[2pt]
\hdashline
\addlinespace[2pt]
\multirow{3}{*}{Qwen3.5-9B~\cite{yang2025qwen3}}
              & \ding{55} & \ding{55}                            & 1.85\,$\pm$\,0.19            & 1.65\,$\pm$\,0.17            & 1.73\,$\pm$\,0.09            & 1.74 \\
              & \ding{55} & \textbf{\ding{51}}                   & \textbf{3.60\,$\pm$\,0.16}   & \textbf{2.99\,$\pm$\,0.19}   & \textbf{2.88\,$\pm$\,0.09}   & \textbf{3.16} \\
              & \multicolumn{2}{c}{$\Delta$\;\textcolor{black!50}{\scriptsize(+Gen. ft.)}}                  & \textcolor{teal}{+1.75}      & \textcolor{teal}{+1.34}      & \textcolor{teal}{+1.15}      & \textcolor{teal}{+1.41}      \\
\cmidrule(lr){1-7}
\multirow{3}{*}{InternVL3.5-8B~\cite{wang2025internvl35}}
              & \ding{55} & \ding{55}                            & 1.79\,$\pm$\,0.18            & 1.43\,$\pm$\,0.121           & 1.56\,$\pm$\,0.09            & 1.59 \\
              & \ding{55} & \textbf{\ding{51}}                   & \textbf{3.60\,$\pm$\,0.17}   & \textbf{3.24\,$\pm$\,0.18}   & \textbf{2.79\,$\pm$\,0.09}   & \textbf{3.21} \\
              & \multicolumn{2}{c}{$\Delta$\;\textcolor{black!50}{\scriptsize(+Gen. ft.)}}                  & \textcolor{teal}{+1.81}      & \textcolor{teal}{+1.81}      & \textcolor{teal}{+1.23}      & \textcolor{teal}{+1.62}      \\
\cmidrule(lr){1-7}
\multirow{3}{*}{Qwen3-VL-8B~\cite{bai2025qwen3vl}}
              & \ding{55} & \ding{55}                            & 1.93\,$\pm$\,0.11            & 1.57\,$\pm$\,0.10            & 1.68\,$\pm$\,0.08            & 1.73 \\
              & \ding{55} & \textbf{\ding{51}}                   & \textbf{2.96\,$\pm$\,0.16}   & \textbf{2.50\,$\pm$\,0.15}   & \textbf{2.25\,$\pm$\,0.10}   & \textbf{2.57} \\
              & \multicolumn{2}{c}{$\Delta$\;\textcolor{black!50}{\scriptsize(+Gen. ft.)}}                  & \textcolor{teal}{+1.03}      & \textcolor{teal}{+0.93}      & \textcolor{teal}{+0.57}      & \textcolor{teal}{+0.84}      \\
\midrule
\multicolumn{7}{l}{\textit{(b) Benefits of fine-tuning the Interaction Planner}} \\
\multirow{3}{*}{Qwen3.5-9B~\cite{yang2025qwen3}}
              & \ding{55} & \ding{51}                            & 3.60\,$\pm$\,0.16            & 2.99\,$\pm$\,0.19            & 2.88\,$\pm$\,0.09            & 3.16 \\
              & \textbf{\ding{51}} & \textbf{\ding{51}}           & \textbf{3.91\,$\pm$\,0.16}   & \textbf{3.63\,$\pm$\,0.19}   & \textbf{3.25\,$\pm$\,0.09}   & \textbf{3.60} \\
              & \multicolumn{2}{c}{$\Delta$\;\textcolor{black!50}{\scriptsize(+Plan. ft.)}}                & \textcolor{teal}{+0.31}      & \textcolor{teal}{+0.64}      & \textcolor{teal}{+0.37}      & \textcolor{teal}{+0.44}      \\
\cmidrule(lr){1-7}
\multirow{3}{*}{InternVL3.5-8B~\cite{wang2025internvl35}}
              & \ding{55} & \ding{51}                            & 3.60\,$\pm$\,0.17            & 3.24\,$\pm$\,0.18            & 2.79\,$\pm$\,0.09            & 3.21 \\
              & \textbf{\ding{51}} & \textbf{\ding{51}}           & \textbf{3.89\,$\pm$\,0.10}   & \textbf{3.36\,$\pm$\,0.16}   & \textbf{2.89\,$\pm$\,0.09}   & \textbf{3.38} \\
              & \multicolumn{2}{c}{$\Delta$\;\textcolor{black!50}{\scriptsize(+Plan. ft.)}}                & \textcolor{teal}{+0.29}      & \textcolor{teal}{+0.12}      & \textcolor{teal}{+0.10}      & \textcolor{teal}{+0.17}      \\
\cmidrule(lr){1-7}
\multirow{3}{*}{Qwen3-VL-8B~\cite{bai2025qwen3vl}}
              & \ding{55} & \ding{51}                            & 2.96\,$\pm$\,0.16            & 2.50\,$\pm$\,0.15            & 2.25\,$\pm$\,0.10            & 2.57 \\
              & \textbf{\ding{51}} & \textbf{\ding{51}}           & \textbf{3.39\,$\pm$\,0.13}   & \textbf{3.14\,$\pm$\,0.13}   & \textbf{2.71\,$\pm$\,0.10}   & \textbf{3.08} \\
              & \multicolumn{2}{c}{$\Delta$\;\textcolor{black!50}{\scriptsize(+Plan. ft.)}}                & \textcolor{teal}{+0.43}      & \textcolor{teal}{+0.64}      & \textcolor{teal}{+0.46}      & \textcolor{teal}{+0.51}      \\
\bottomrule
\end{tabular}

\end{adjustbox}
\end{table}

\Needspace{6\baselineskip}
\subsection{Evaluation Set Construction}
\label{app:eval_set}

We construct the held-out evaluation set from the real-query subset, because genuine live-chat triggers remove any dependence of the benchmark on reconstruction quality. The final set contains \textbf{100} samples, selected to cover the four major video categories in balanced proportion as well as the range of query types, response durations, human counts, and interaction patterns present in the corpus.

\noindent\textbf{Split disjointness.} Training, validation, and test splits are disjoint at both the \emph{channel and streamer} levels. No channel or streamer represented in the evaluation set appears in training or validation, including when a streamer appears on multiple channels. This is stricter than clip-level exclusion and matters here because a single livestream channel yields many clips with the same streamer, background, microphone, and verbal habits; with clip-level splits alone, a model could score well by reproducing a memorized identity rather than by responding to the query. We additionally deduplicate near-identical clips within a channel before assigning splits.

\noindent\textbf{Why 100.} The evaluation size is limited by inference cost, not by data availability. Generating one response takes approximately six minutes on eight H100 GPUs, so a single configuration on 100 videos costs about 10 wall-clock hours or roughly 80 H100 GPU-hours; the full grid reported in the paper therefore costs several hundred GPU-hours.

\Needspace{6\baselineskip}
\subsection{Statistical Reporting}
\label{app:statistical_reporting}

Given scores $s_{m,k}$ for model $m$ on sample $k$ over $K$ samples, we report the mean score
\begin{equation}
    \bar{s}_m = \frac{1}{K}\sum_{k=1}^{K}s_{m,k},
\end{equation}
and quantify uncertainty with the standard error of the mean,
\begin{equation}
    \mathrm{SE}_m
    =
    \frac{\sigma_m}{\sqrt{K}},
    \qquad
    \sigma_m =
    \sqrt{
    \frac{1}{K-1}
    \sum_{k=1}^{K}(s_{m,k}-\bar{s}_m)^2
    }.
\end{equation}
All $\pm$ values in \Cref{tab:main} are standard errors, not standard deviations. This is worth stating explicitly because the two differ by a factor of $\sqrt{K}=10$ at $K=100$: the standard error describes uncertainty in the \emph{estimated mean}, which is the quantity being compared across systems, whereas the standard deviation describes the dispersion of individual ratings and is much larger.

For paired comparisons, we compute per-sample differences between two systems, which removes per-example difficulty as a source of variance, and report the mean difference. Because the ratings are ordinal and bounded, we assess significance with a paired bootstrap over samples ($10^{4}$ resamples) rather than assuming normality.

\Needspace{6\baselineskip}
\subsection{Human Ratings Against the Automatic Judge}
\label{app:human_contrasts}

\Cref{tab:main} combines VLM-judge scores on all 100 held-out examples with human scores on a sampled subset. Here, \Cref{tab:human_vs_judge} compares both evaluators on exactly the same generated videos, using ten videos per configuration. Its human columns are identical to those in \Cref{tab:main}, whereas its judge columns use only the human-evaluation subset. We report Pearson and Spearman correlations between configuration-level means over the eight configurations.

\begin{table}[t]
\centering
\small
\setlength{\tabcolsep}{4pt}
\renewcommand{\arraystretch}{1.05}
\caption{Human (H) and VLM-judge (J) scores on matched videos. OVRL averages three semantic dimensions; video/audio quality is human-rated. Correlations use the eight configuration means.}
\label{tab:human_vs_judge}
\begin{adjustbox}{max width=\linewidth}
\begin{tabular}{l cc cc cc c c cc}
\toprule
 & \multicolumn{2}{c}{\cellcolor{green!30}\textsc{Relev.}\,$\uparrow$} & \multicolumn{2}{c}{\cellcolor{green!30}\textsc{Approp.}\,$\uparrow$} & \multicolumn{2}{c}{\cellcolor{green!30}\textsc{Expr.}\,$\uparrow$} & \cellcolor{red!25}\textsc{Vid.}\,$\uparrow$ & \cellcolor{yellow!30}\textsc{Aud.}\,$\uparrow$ & \multicolumn{2}{c}{\cellcolor{green!30}\textsc{OVRL}\,$\uparrow$} \\
\cmidrule(lr){2-3} \cmidrule(lr){4-5} \cmidrule(lr){6-7} \cmidrule(lr){8-8} \cmidrule(lr){9-9} \cmidrule(lr){10-11}
Planner + Generator & H & J & H & J & H & J & H & H & H & J \\
\midrule
Oracle (GT) + MOVA \emph{(pt.)}      & 4.33 & 4.93 & 3.89 & 4.64 & 3.72 & 3.29 & 3.17 & 3.61 & 3.98 & 4.29 \\
Oracle (GT) + MOVA \emph{(ft.)}      & \textbf{4.41} & \textbf{5.00} & \textbf{3.98} & \textbf{4.67} & \textbf{3.96} & \textbf{3.50} & \textbf{4.00} & \textbf{3.98} & \textbf{4.12} & \textbf{4.39} \\
\midrule
Qwen3.5 \emph{(pt.)} + MOVA \emph{(pt.)}     & 2.62 & 1.82 & 1.92 & 1.46 & 2.22 & 1.57 & 1.93 & 2.48 & 2.25 & 1.62 \\
Qwen3.5 \emph{(pt.)} + MOVA \emph{(ft.)}     & 3.55 & 3.71 & 3.14 & 3.64 & 3.17 & 3.00 & 3.53 & 3.79 & 3.29 & 3.45 \\
Qwen3.5 \emph{(ft.)} + MOVA \emph{(ft.)}     & 4.00 & 4.11 & 3.59 & 3.86 & 3.63 & 3.32 & 3.81 & 3.96 & 3.74 & 3.76 \\
\midrule
InternVL3.5 \emph{(pt.)} + MOVA \emph{(pt.)} & 2.04 & 1.86 & 1.63 & 1.32 & 1.48 & 1.57 & 1.37 & 1.81 & 1.72 & 1.58 \\
InternVL3.5 \emph{(pt.)} + MOVA \emph{(ft.)} & 3.86 & 3.50 & 3.08 & 3.21 & 3.26 & 2.79 & 3.86 & 3.84 & 3.40 & 3.17 \\
InternVL3.5 \emph{(ft.)} + MOVA \emph{(ft.)} & 3.96 & 3.61 & 3.48 & 3.32 & 3.63 & 2.82 & 3.63 & 3.78 & 3.69 & 3.25 \\
\midrule
Pearson $r$      & \multicolumn{2}{c}{0.949} & \multicolumn{2}{c}{0.979} & \multicolumn{2}{c}{0.947} & -- & -- & \multicolumn{2}{c}{\textbf{0.964}} \\
Spearman $\rho$  & \multicolumn{2}{c}{0.905} & \multicolumn{2}{c}{0.976} & \multicolumn{2}{c}{0.867} & -- & -- & \multicolumn{2}{c}{\textbf{0.929}} \\
\bottomrule
\end{tabular}
\end{adjustbox}
\end{table}

Every contrast underlying our conclusions agrees in sign between the two protocols, and the ordering of magnitudes is preserved: generator fine-tuning is the largest effect ($+1.68$ and $+1.04$ OVRL by human rating, against $+1.59$ and $+1.83$ by the judge), planner fine-tuning is a smaller positive effect ($+0.29$ and $+0.45$, against $+0.08$ and $+0.31$), and both learned planners remain behind oracle captions ($-0.43$ and $-0.38$, against $-1.14$ and $-0.63$). The two scales differ in a systematic way: the judge is harsher than humans at the bottom of the range, it compresses the expressiveness scale, and it rates oracle-caption generations closer to the ceiling than human raters do, which widens the oracle-to-planner gap on the judge scale. Human-rated video quality tracks generator fine-tuning strongly and is, correctly, almost unaffected by planner fine-tuning, since the planner only alters the conditioning text.

\Needspace{6\baselineskip}
\subsection{Counterfactual Query Evaluation}
\label{app:counterfactual}

A model could score well on our benchmark by producing a generic, context-appropriate livestream reaction while ignoring the query entirely. Because relevance is scored by a judge that also sees the context, a purely scene-driven system would not necessarily be penalized as heavily as it should be. To rule this out, we sampled 20 evaluation examples, replaced each query with a different but still plausible livestream query while holding the visual context fixed, reran inference with the fine-tuned Qwen3.5 planner and fine-tuned MOVA generator, and scored the outputs against the \emph{substituted} query. \Cref{tab:counterfactual} reports the result.

\begin{table}[t]
\centering
\small
\setlength{\tabcolsep}{8pt}
\renewcommand{\arraystretch}{1.1}
\caption{Counterfactual query evaluation on 20 examples with the fine-tuned Qwen3.5 planner and fine-tuned MOVA generator. In the counterfactual condition the query is replaced while the context is held fixed, and outputs are scored against the \emph{substituted} query.}
\label{tab:counterfactual}
\begin{adjustbox}{max width=\linewidth}
\begin{tabular}{lcccc}
\toprule
Setting & Relev.\,$\uparrow$ & Approp.\,$\uparrow$ & Expr.\,$\uparrow$ & OVRL\,$\uparrow$ \\
\midrule
Original query        & 4.03 & 3.82 & 3.35 & 3.73 \\
Counterfactual query  & 3.98 & 3.91 & 3.28 & 3.72 \\
\midrule
$\Delta$              & $-$0.05 & +0.09 & $-$0.07 & $-$0.01 \\
\bottomrule
\end{tabular}
\end{adjustbox}
\end{table}

Scores are effectively unchanged. This is the outcome that distinguishes query-following from context-driven generation: a model that ignored the query and produced a plausible reaction from the visual context alone would retain its appropriateness score but lose relevance to the substituted query, since the substituted query asks about something else. Relevance instead stays at $3.98$ against $4.03$, well inside sampling noise. Qualitatively, the change in behavior is grounded in the new query rather than lexical: asked ``do you have a pattern for the cardigan in the back?'' over an unchanged crafting scene, the generated streamer lifts the cardigan and turns its pattern towards the camera, an action absent from the response to the original query. Within plausible livestream queries, the response is therefore driven by the query and not only by the scene.

\Needspace{6\baselineskip}
\subsection{Two-Stage Pipeline Audio-Visual Quality}
\label{app:two_stage_av}

\Cref{tab:av_eval} conditions the generator on oracle captions, which isolates generator adaptation but leaves open whether the low-level gains survive when the conditioning text is produced by a real planner. \Cref{tab:two_stage_av} therefore repeats the audio-visual metrics for the full two-stage pipeline, sweeping pretrained and fine-tuned Qwen3-VL and InternVL-3.5 planners against pretrained and fine-tuned MOVA.

\begin{table}[t]
\centering
\small
\setlength{\tabcolsep}{4pt}
\renewcommand{\arraystretch}{1.1}
\caption{Audio-visual quality of the complete two-stage pipeline, with planner outputs rather than oracle captions as the conditioning text. Q = Qwen3-VL, I = InternVL-3.5; \emph{pt} and \emph{ft} denote pretrained and fine-tuned on InteracVid. \textbf{Bold} marks the best value per metric.}
\label{tab:two_stage_av}
\begin{adjustbox}{max width=\linewidth}
\begin{tabular}{lcccc?ccc?ccc}
\toprule
 & \multicolumn{4}{c}{\cellcolor{red!30}\textsc{Video quality}} & \multicolumn{3}{c}{\cellcolor{yellow!30}\textsc{Audio quality}} & \multicolumn{3}{c}{\cellcolor{blue!30}\textsc{A--V sync.}} \\
\cmidrule(lr){2-5} \cmidrule(lr){6-8} \cmidrule(lr){9-11}
Pipeline & ID\,$\uparrow$ & SC\,$\uparrow$ & BC\,$\uparrow$ & MS\,$\downarrow$ & CE\,$\uparrow$ & CU\,$\uparrow$ & PQ\,$\uparrow$ & LC\,$\uparrow$ & LD\,$\downarrow$ & AVA\,$\uparrow$ \\
\midrule
Q\,\emph{pt} + MOVA\,\emph{pt} & 0.8293 & 0.9636 & 0.9594 & 0.0113 & 4.779 & 5.084 & 5.933 & 1.358 & 9.382 & 0.2625 \\
Q\,\emph{pt} + MOVA\,\emph{ft} & 0.8609 & \textbf{0.9731} & 0.9598 & \textbf{0.0078} & 5.153 & 5.737 & 6.356 & 1.456 & 9.301 & 0.2781 \\
Q\,\emph{ft} + MOVA\,\emph{ft} & \textbf{0.8709} & 0.9689 & \textbf{0.9599} & 0.0095 & \textbf{5.273} & 5.946 & 6.490 & \textbf{1.488} & 9.180 & \textbf{0.2830} \\
\midrule
I\,\emph{pt} + MOVA\,\emph{pt} & 0.8095 & 0.9582 & 0.9461 & 0.0131 & 4.708 & 5.044 & 5.960 & 1.136 & 9.199 & 0.2099 \\
I\,\emph{pt} + MOVA\,\emph{ft} & 0.8531 & 0.9677 & 0.9546 & 0.0095 & 5.126 & 5.722 & 6.366 & 1.370 & 9.117 & 0.2576 \\
I\,\emph{ft} + MOVA\,\emph{ft} & 0.8671 & 0.9663 & 0.9573 & 0.0104 & 5.259 & \textbf{5.938} & \textbf{6.495} & 1.444 & \textbf{9.088} & 0.2728 \\
\bottomrule
\end{tabular}
\end{adjustbox}
\end{table}

The pattern from the oracle-caption setting carries over. For both planner backbones, fine-tuning the generator improves every metric, and fine-tuning the planner on top of that improves identity preservation, all three audio-quality metrics, lip-sync distance, and AV-Align. The exceptions are informative: subject consistency and motion smoothness peak at the \emph{pt}~planner + \emph{ft}~generator configuration rather than at the fully fine-tuned one. This is not a regression in quality but a consequence of what the planner learns. A fine-tuned planner writes richer, more specific action descriptions, which the generator renders as more motion; more motion mechanically lowers frame-to-frame consistency and smoothness scores, both of which are maximized by near-static video. The same effect appears in \Cref{tab:av_eval_full}, where dynamic degree is the one temporal metric that our fine-tuned generator does not improve. Read together with the semantic results in \Cref{tab:main}, this is the intended trade: the pipeline moves away from static talking-head output towards expressive reactions.

\Needspace{6\baselineskip}
\subsection{Effect of Preceding Audio-Visual Context}
\label{app:context_ablation}

The last two rows of \Cref{tab:av_eval}~(a) compare LTX-2.3~\cite{lightricks2026ltx2} conditioned on the first frame alone against the same model additionally conditioned on the preceding audio and video. The gains there are concentrated in audio: content enjoyment, content usefulness, and production quality all improve substantially, and lip-sync distance drops by $0.84$, the single largest change in the comparison. Visual metrics move by less than $0.006$ in either direction, which is what we expect, since the first frame already fixes identity, framing, and background, and none of these metrics measures motion continuity with the preceding shot --- precisely the property that temporal context should improve. Two metrics omitted from the main table behave consistently with this reading: dynamic degree rises slightly ($1.926 \rightarrow 1.968$) and the aesthetic score is unchanged ($0.3115 \rightarrow 0.3109$). AV-Align decreases, which we attribute to the richer audio track produced under full context being harder to align against onset-based visual events.

To confirm that the audio gains reflect genuine acoustic conditioning rather than a scoring artifact, we probe speaker identity directly with ECAPA-TDNN~\cite{desplanques2020ecapatdnn} embeddings, comparing generated speech against the real speaker from the same stream. \Cref{tab:speaker_sim} reports cosine similarity with two references: a \emph{mismatch floor}, computed between speakers from different streams, and a \emph{real--real ceiling}, computed between two real segments of the same speaker.

\begin{table}[t]
\centering
\small
\setlength{\tabcolsep}{8pt}
\renewcommand{\arraystretch}{1.1}
\caption{ECAPA-TDNN speaker similarity between generated and reference speech. The floor and ceiling rows calibrate the scale: the floor is the similarity between unrelated speakers, and the ceiling is between two real segments of the same speaker.}
\label{tab:speaker_sim}
\begin{adjustbox}{max width=\linewidth}
\begin{tabular}{lcc}
\toprule
Setting & Cosine similarity & 95\% CI \\
\midrule
Generated, without preceding audio & 0.1156 & $\pm$0.0264 \\
Generated, with preceding audio    & \textbf{0.2747} & $\pm$0.0479 \\
\midrule
Mismatch floor (different speakers) & 0.0326 & $\pm$0.0091 \\
Real--real ceiling (same speaker)   & 0.4241 & $\pm$0.0460 \\
\bottomrule
\end{tabular}
\end{adjustbox}
\end{table}

Supplying preceding audio more than doubles speaker similarity, from $0.116$ to $0.275$, and the confidence intervals do not overlap. Interpreted against the calibration rows, first-frame conditioning recovers about $21\%$ of the floor-to-ceiling range while full-context conditioning recovers about $62\%$. Temporal audio therefore carries speaker identity and acoustic state that no static frame can encode, which is exactly the information InteracVid's preceding context preserves and that current first-frame interfaces discard.

\Needspace{6\baselineskip}
\subsection{Full Evaluation Results on the AV Co-Generator}
\label{app:full_metrics}

Because of space constraints, the main text reports only a subset of our audio-visual metrics. \Cref{tab:av_eval_full} lists the complete set of 26 quantitative metrics assessing generation quality.

\noindent\textbf{Metrics in the main comparison.} Video metrics are Identity Score (ID) from Verse-Bench~\cite{wang2025universe}, and Subject Consistency (SC), Background Consistency (BC), Motion Smoothness (MS), and Temporal Flickering (TF) from VBench~\cite{huang2024vbench}. Audio metrics are Content Enjoyment (CE), Content Usefulness (CU), and Production Quality (PQ) from AudioBox-Aesthetics~\cite{tjandra2025audiobox}, together with Character Error Rate (CER). For audio--video synchronization, we report LipSync Confidence (LC), LipSync Distance (LD), and AV-Align (AVA) from Verse-Bench.

\begin{table}[t]
\centering
\small
\setlength{\tabcolsep}{5pt}
\renewcommand{\arraystretch}{1.07}
\caption{Full 26-metric evaluation. Ours is MOVA fine-tuned on InteracVid; bold marks gains over pretrained MOVA. Positive $\Delta$ denotes improvement in native metric units; $\circ$ marks signed-offset diagnostics without a preferred direction.}
\label{tab:av_eval_full}
\begin{adjustbox}{max width=\linewidth}
\begin{tabular}{l c ccc ? cc ? c}
\toprule
 & & \multicolumn{3}{c}{\textit{Zero-shot baselines}} & \multicolumn{2}{c}{\textit{Main comparison}} & \\
\cmidrule(lr){3-5} \cmidrule(lr){6-7}
\textbf{Metric} & \textbf{Dir.} & OVI & UniAVGen & UniVerse-1 & MOVA & \textbf{Ours} & $\Delta$ \\
\midrule
\multicolumn{8}{l}{\textsc{Visual quality}} \\
\addlinespace[1pt]
AS\textsubscript{LAION+MUSIQ+ManIQA} & $\uparrow$ & 0.320 & 0.282 & 0.385 & 0.369 & \textbf{0.373} & \textcolor{black!60}{\scriptsize+0.004} \\
AS-LAION\textsubscript{aes-pred} & $\uparrow$ & 0.151 & 0.151 & 0.156 & 0.153 & 0.153 & \textcolor{black!60}{\scriptsize$-$0.001} \\
MUSIQ & $\uparrow$ & 0.300 & 0.189 & 0.492 & 0.446 & \textbf{0.459} & \textcolor{black!60}{\scriptsize+0.012} \\
ManIQA & $\uparrow$ & 0.508 & 0.508 & 0.508 & 0.508 & \textbf{0.508} & \textcolor{black!60}{\scriptsize+0.000} \\
ID\textsubscript{DINOv2} & $\uparrow$ & 0.905 & 0.914 & 0.911 & 0.868 & \textbf{0.881} & \textcolor{black!60}{\scriptsize+0.013} \\
\midrule
\multicolumn{8}{l}{\textsc{Temporal coherence}} \\
\addlinespace[1pt]
Subject consistency & $\uparrow$ & 0.971 & 0.971 & 0.986 & 0.950 & \textbf{0.962} & \textcolor{black!60}{\scriptsize+0.012} \\
Background consistency & $\uparrow$ & 0.969 & 0.973 & 0.976 & 0.949 & \textbf{0.960} & \textcolor{black!60}{\scriptsize+0.011} \\
Motion smoothness & $\downarrow$ & 0.008 & 0.013 & 0.004 & 0.023 & \textbf{0.014} & \textcolor{black!60}{\scriptsize+0.009} \\
Temporal flicker & $\downarrow$ & 0.011 & 0.016 & 0.005 & 0.029 & \textbf{0.019} & \textcolor{black!60}{\scriptsize+0.010} \\
Dynamic degree & $\uparrow$ & 1.347 & 1.668 & 0.444 & 2.369 & 2.178 & \textcolor{black!60}{\scriptsize$-$0.191} \\
\midrule
\multicolumn{8}{l}{\textsc{Cross-modal alignment}} \\
\addlinespace[1pt]
CLIP-T\textsubscript{ViT-L/14} & $\uparrow$ & 0.257 & 0.256 & 0.255 & 0.249 & \textbf{0.256} & \textcolor{black!60}{\scriptsize+0.007} \\
\midrule
\multicolumn{8}{l}{\textsc{Audio quality}} \\
\addlinespace[1pt]
CE\textsubscript{AudioBox} (Content Enj.) & $\uparrow$ & 5.112 & 5.242 & 3.413 & 5.217 & \textbf{5.392} & \textcolor{black!60}{\scriptsize+0.175} \\
CU\textsubscript{AudioBox} (Content Usef.) & $\uparrow$ & 5.576 & 5.811 & 4.352 & 5.823 & \textbf{6.154} & \textcolor{black!60}{\scriptsize+0.331} \\
PC\textsubscript{AudioBox} (Prod. Compl.) & $\downarrow$ & 1.756 & 2.210 & 2.196 & 1.667 & 1.756 & \textcolor{black!60}{\scriptsize$-$0.088} \\
PQ\textsubscript{AudioBox} (Prod. Qual.) & $\uparrow$ & 6.002 & 6.518 & 4.879 & 6.273 & \textbf{6.625} & \textcolor{black!60}{\scriptsize+0.352} \\
DNSMOS\textsubscript{P.835} & $\uparrow$ & 3.222 & 2.977 & 2.197 & 2.900 & \textbf{2.958} & \textcolor{black!60}{\scriptsize+0.059} \\
\midrule
\multicolumn{8}{l}{\textsc{Speech intelligibility}} \\
\addlinespace[1pt]
WER\textsubscript{Whisper-tiny} & $\downarrow$ & 0.297 & 0.796 & 0.649 & 0.236 & \textbf{0.222} & \textcolor{black!60}{\scriptsize+0.014} \\
CER\textsubscript{Whisper-tiny} & $\downarrow$ & 0.251 & 0.694 & 0.596 & 0.193 & \textbf{0.179} & \textcolor{black!60}{\scriptsize+0.014} \\
\midrule
\multicolumn{8}{l}{\textsc{Audio--visual synchronisation}} \\
\addlinespace[1pt]
Lip-sync conf.\textsubscript{SyncNet} & $\uparrow$ & 1.984 & 1.403 & 0.354 & 1.456 & \textbf{1.519} & \textcolor{black!60}{\scriptsize+0.063} \\
Lip-sync dist.\textsubscript{SyncNet} & $\downarrow$ & 8.483 & 9.868 & 10.688 & 10.062 & \textbf{9.651} & \textcolor{black!60}{\scriptsize+0.411} \\
Lip-sync offset\textsubscript{SyncNet} (frames) & $\circ$ & $-$1.947 & 1.050 & 0.818 & $-$1.000 & $-$4.200 & -- \\
AV-Align\textsubscript{onset-IoU} & $\uparrow$ & 0.240 & 0.316 & 0.304 & 0.248 & \textbf{0.288} & \textcolor{black!60}{\scriptsize+0.040} \\
ImageBind cosine\textsubscript{ImageBind-Huge} & $\uparrow$ & 0.253 & 0.201 & 0.145 & 0.245 & 0.230 & \textcolor{black!60}{\scriptsize$-$0.015} \\
DeSync\textsubscript{ImageBind} (s) & $\downarrow$ & 0.650 & 0.600 & 0.679 & 0.664 & \textbf{0.636} & \textcolor{black!60}{\scriptsize+0.029} \\
DeSync signed offset\textsubscript{ImageBind} (s) & $\circ$ & 0.293 & 0.157 & $-$0.036 & $-$0.221 & $-$0.264 & -- \\
DeSync peak sim.\textsubscript{ImageBind} & $\uparrow$ & 0.289 & 0.218 & 0.165 & 0.273 & 0.259 & \textcolor{black!60}{\scriptsize$-$0.014} \\
\bottomrule
\end{tabular}
\end{adjustbox}
\end{table}

Dynamic degree is the one temporal metric on which fine-tuning does not help, dropping from $2.369$ to $2.178$. We read this as a consequence of domain adaptation rather than a loss of quality: the pretrained model produces looser, less controlled motion, whereas the fine-tuned model matches the tighter framing of livestream footage. The same tension appears in \Cref{tab:two_stage_av}, where consistency and smoothness are highest for the configuration that generates the least motion.

\section{LLM and VLM Prompts}
\label{app:prompts}

This supplementary section lists the prompts used in the curation and evaluation pipeline. Together with the fixed model versions reported in \Cref{app:llm_usage}, they document the automatic evaluation protocol; \Cref{app:human_contrasts} compares the resulting scores with human ratings.

\Needspace{6\baselineskip}
\subsection{Subtitle Normalization Prompt}

\begin{lstlisting}[style=promptstyle, caption={Prompt used for subtitle normalization}, label={lst:prompt_subtitle}]
You are an expert AI transcription formatter. Your task is to convert raw SRT subtitle data into a structured JSON transcript.

**Input Data:**
SRT_DATA:
"""
{srt}
"""

**Instructions:**

1.  **Sentence Reconstruction (Strict Verbatim):**
    * **Merge & Split:** Combine fragmented subtitle lines into complete sentences based on punctuation and context.
    * **Strictly Verbatim:** Do NOT remove filler words (e.g., "Um", "uh"), stammers, or false starts. Preserve the text exactly as spoken.
    * **No Paraphrasing:** Do not correct grammar or vocabulary. Keep the original wording entirely intact.

2.  **Timestamp Logic (Block Approximation):**
    * **Start Time:** Use the `start` timestamp of the *first* subtitle block that contains any part of this sentence.
    * **End Time:** Use the `end` timestamp of the *last* subtitle block that contains any part of this sentence.
    * *Note:* If a subtitle block contains the boundary between two sentences (e.g., "End of sentence A. Start of sentence B"), both sentences will share that block's timestamp range. This is an expected rough approximation.

3.  **Formatting:**
    * Output strictly valid JSON.
    * Schema: `[{"start": "string", "end": "string", "text": "string"}]`

**Example:**

*Input:*
1
00:00:01,000 --> 00:00:05,000
I ate the apple. Then I

2
00:00:05,000 --> 00:00:08,000
went to sleep. Um, yeah.

*Output Logic:*
* "I ate the apple." is entirely in Block 1. -> Start: 00:00:01,000 | End: 00:00:05,000
* "Then I went to sleep." starts in Block 1 and ends in Block 2. -> Start: 00:00:01,000 | End: 00:00:08,000
* "Um, yeah." is entirely in Block 2. -> Start: 00:00:05,000 | End: 00:00:08,000

*Output JSON:*
[
  {
    "start": "00:00:01,000",
    "end": "00:00:05,000",
    "text": "I ate the apple."
  },
  {
    "start": "00:00:01,000",
    "end": "00:00:08,000",
    "text": "Then I went to sleep."
  },
  {
    "start": "00:00:05,000",
    "end": "00:00:08,000",
    "text": "Um, yeah."
  }
]

**Task:**
Process the provided SRT_DATA into the JSON format. Return the JSON object only.
\end{lstlisting}

\Needspace{6\baselineskip}
\subsection{Real Comment--Response Matching Prompt}

\begin{lstlisting}[style=promptstyle, caption={Prompt used for live-chat query-response matching}, label={lst:prompt_match}]
JSON_DATA:
"""{json}"""

You are an expert AI data annotator specializing in live stream interaction analysis.
You will receive a JSON list of "speech events." Each event contains a sentence spoken by a streamer (`target`), their surrounding context, and a list of potential chat triggers (`candidates`).

For each object, you must:

1. **Classify** the `target` sentence into one of three categories: `continuation`, `reactive_response`, or `proactive_monologue`.
2. **Identify** the specific chat message (bullet) that triggered the response, if the sentence is classified as `reactive_response`.

### Input Data Structure

* `target`: The main sentence to classify.
* `context_prev`: The sentence spoken immediately before.
* `candidates`: A list of chat messages. Each has a unique `bullet_index`.

---

### Classification Logic (Apply to EACH item)

**1. CONTINUATION** (`continuation`)
* **Definition:** The `target` is grammatically or semantically "glued" to `context_prev`. The streamer is finishing a thought, sentence, or story started in the previous segment.
* **Indicators:**
  * Starts with conjunctions (And, But, So, Because).
  * Contains pronouns refer to nouns in `context_prev` (e.g., Prev: "I bought a dress." -> Target: "**It** fits really well.")
  * Lacks new address to a new content or a user.
* **Decision:** If `target` simply continues the previous sentence, label `continuation`.

**2. REACTIVE RESPONSE** (`reactive_response`)
* **Definition:** The `target` is the **start** of a direct response to a specific chat message in `candidates`.
* **Indicators:**
    * **Name Match:** Says a username found in a candidate. (e.g., "Hi **Nina**!").
    * **Semantic Match:** Answers a question or comments on a specific topic in a candidate. (e.g., Candidate: "Is that wool?" -> Target: "Yes, this is 100% wool.").
    * **Repetition:** Repetes one of the messages followed by further comments or responses. (e.g., Candidate: "Too loud" -> Target: "Too loud, oh I will turn down the volumn").
    * **Greeting:** Welcomes a specific user.
* **Constraint:** The candidate must logically prompt the speech.
* **Decision:** Label `reactive_response` and output the `trigger_bullet_index`.

**3. PROACTIVE MONOLOGUE** (`proactive_monologue`)
* **Definition:** The streamer starts a new topic, story, or instructional segment that is **not** triggered by any `candidates` and is **not** a continuation.
* **Indicators:**
  * Topic shifts (e.g., "Anyway, moving on to the next item...").
  * General statements to the whole audience, not a specific person.
  * Self-correction or technical adjustments (e.g., "Let me fix the camera").
* **Decision:** If it fits neither of the above, label `proactive_monologue`.

---

### Output Format

Return the exact input JSON array, but add three fields to every object:

1. `label`: The classification (`continuation`, `reactive_response`, or `proactive_monologue`).
2. `trigger_bullet_index`: The `bullet_index` of the candidate that triggered the response. **Return `null**` if the label is not `reactive_response`.
3. `reason`: A brief explanation of why you chose the label and (if applicable) why you selected that specific candidate.

**Example Output:**

```json
[
    {
        "meta": { ... },
        "target": { "text": "Hi Nina, glad you are here!" ... },
        "context_prev": "I just started.",
        "candidates": [ ... ],
        "label": "reactive_response",
        "trigger_bullet_index": 0,
        "reason": "Streamer explicitly greets user 'ninaallen4206' who commented 'Hello' in the candidate list."
    }
]
```
\end{lstlisting}

\Needspace{6\baselineskip}
\subsection{Discourse Role Classification Prompt}

\begin{lstlisting}[style=promptstyle, caption={Prompt used for discourse role classification of transcript without live-chat metadata}, label={lst:prompt_discourse}]
JSON_DATA:
"""{json}"""

You are an expert AI data annotator specializing in live stream interaction analysis.
You will receive a JSON list of "speech events." Each event contains a sentence spoken by a streamer (`target`), their surrounding context, and a list of potential chat triggers (`candidates`).

For each object, you must:

1. **Classify** the `target` sentence into one of three categories: `continuation`, `reactive_response`, or `proactive_monologue`.
2. **Identify** the specific chat message (bullet) that triggered the response, if the sentence is classified as `reactive_response`.

### Input Data Structure

* `target`: The main sentence to classify.
* `context_prev`: The sentence spoken immediately before.
* `candidates`: A list of chat messages. Each has a unique `bullet_index`.

---

### Classification Logic (Apply to EACH item)

**1. CONTINUATION** (`continuation`)
* **Definition:** The `target` is grammatically or semantically "glued" to `context_prev`. The streamer is finishing a thought, sentence, or story started in the previous segment.
* **Indicators:**
  * Starts with conjunctions (And, But, So, Because).
  * Contains pronouns refer to nouns in `context_prev` (e.g., Prev: "I bought a dress." -> Target: "**It** fits really well.")
  * Lacks new address to a new content or a user.
* **Decision:** If `target` simply continues the previous sentence, label `continuation`.

**2. REACTIVE RESPONSE** (`reactive_response`)
* **Definition:** The `target` is the **start** of a direct response to a specific chat message in `candidates`.
* **Indicators:**
    * **Name Match:** Says a username found in a candidate. (e.g., "Hi **Nina**!").
    * **Semantic Match:** Answers a question or comments on a specific topic in a candidate. (e.g., Candidate: "Is that wool?" -> Target: "Yes, this is 100% wool.").
    * **Repetition:** Repetes one of the messages followed by further comments or responses. (e.g., Candidate: "Too loud" -> Target: "Too loud, oh I will turn down the volumn").
    * **Greeting:** Welcomes a specific user.
* **Constraint:** The candidate must logically prompt the speech.
* **Decision:** Label `reactive_response` and output the `trigger_bullet_index`.

**3. PROACTIVE MONOLOGUE** (`proactive_monologue`)
* **Definition:** The streamer starts a new topic, story, or instructional segment that is **not** triggered by any `candidates` and is **not** a continuation.
* **Indicators:**
  * Topic shifts (e.g., "Anyway, moving on to the next item...").
  * General statements to the whole audience, not a specific person.
  * Self-correction or technical adjustments (e.g., "Let me fix the camera").
* **Decision:** If it fits neither of the above, label `proactive_monologue`.

---

### Output Format

Return the exact input JSON array, but add three fields to every object:

1. `label`: The classification (`continuation`, `reactive_response`, or `proactive_monologue`).
2. `trigger_bullet_index`: The `bullet_index` of the candidate that triggered the response. **Return `null**` if the label is not `reactive_response`.
3. `reason`: A brief explanation of why you chose the label and (if applicable) why you selected that specific candidate.

**Example Output:**

```json
[
    {
        "meta": { ... },
        "target": { "text": "Hi Nina, glad you are here!" ... },
        "context_prev": "I just started.",
        "candidates": [ ... ],
        "label": "reactive_response",
        "trigger_bullet_index": 0,
        "reason": "Streamer explicitly greets user 'ninaallen4206' who commented 'Hello' in the candidate list."
    }
]

```
\end{lstlisting}

\Needspace{6\baselineskip}
\subsection{Synthetic Query Reconstruction Prompt}

\begin{lstlisting}[style=promptstyle, caption={Prompt used for reconstructing live-chat queries}, label={lst:prompt_query_recon}]
**Role**: You are an expert multimodal video analyst and context inferencer. I will provide you with a video (or a sequence of video frames) and its ASR transcript. You know that this video is a reactive response - the subject is reacting to, answering, or responding to an unseen comment or question from the live chat (bullet/danmaku).

**Task**: Your goal is to accurately describe the video and use the visual and audio evidence to synthesize the single most likely comment or question that triggered this response.

**Step 1: Visual & Audio Analysis (The Evidence)**

  * Visual Cues: Scan the screen for on-screen chat, bullets, or danmaku. If there is a visible comment that exactly matches the response, use it directly.
  * Verbal Cues (CRITICAL): Carefully analyze the provided ASR transcript. Infer the actual meaning and intention of the speech. Identify which specific part is the response to a query versus a continuation of proactive streaming. Pay attention to where and how the response begins. Quote the transcript directly in your analysis.

**Step 2: Synthesize the Trigger**
Based on your analysis, deduce the exact wording of the comment made to the subject immediately before the response.

  * Authentic Chat Style: Write the query exactly as a real viewer would type it in a live stream chat. Keep it natural, conversational, and concise (e.g., use "What game is this?" instead of "I am inquiring about the title of the video game you are playing").
  * Absolute Certainty: You must output one definitive query. Do NOT output uncertain thoughts, multiple choices, parentheses containing alternatives (e.g., "or", "like"), or meta-descriptions of the query. Commit to the single most plausible trigger.
  * Formatting: Use the format "<Name>: <Query>". The <Name> should be the inferred name of the user leaving the query (e.g., response: "Hi Alex!" -> trigger: "Alex: Hello!"). If no explicit name can be inferred, strictly use the literal string "<Name>" (e.g., response: "It cost me 30 dollars" -> trigger: "<Name>: How much was that shirt?").
  * No Trigger: If there is no question or comment that could plausibly trigger the response, or if the transcript is purely proactive streaming, strictly output "None".

**TRANSCRIPT**:
"""{TRANSCRIPT}"""

**Output Schema**:
You must return a single valid JSON object. Do not include markdown formatting (```json) or prose.

{
  "analysis": "String detailing visual cues from the video and verbal cues from the transcript.",
  "trigger": "String following exactly '<Name>: <Query>' OR 'None'. No parentheses, alternatives, or uncertain language.",
  "reason": "String explaining your reasoning for synthesizing this specific trigger."
}
\end{lstlisting}

\Needspace{6\baselineskip}
\subsection{Visual and Semantic Quality Assessment Prompt}

\begin{lstlisting}[style=promptstyle, caption={Prompt used for assessing the visual and semantic quality of the video clip}, label={lst:prompt_quality}]
You are a precise Video Content Classifier designed for high-throughput metadata extraction. Given the following transcript, and the accompanying video, you will need to extract structural metadata, identify the number of humans, and rate the correlation between the semantic content of the transcript and the video's visual content. Your output must be valid JSON only.

**Input Context:**

* Transcript: """{TRANSCRIPT}"""

**Task Instructions:**

1. **Language Detection:**
* English Transcript: Evaluate the provided transcript text. If the primary language spoken is English, flag as `true`. If it is in another language, or mixed with English not being the primary language, flag as `false`.

2. **Human Enumeration:**
* Please analyze the video and count the number of humans based on the following strict criteria:
* Main Window Only: Only count humans appearing in the primary content area (the main window). Explicitly ignore any humans appearing in a Picture-in-Picture (PiP) frame, webcam overlay, or corner streamer camera.
* Face Visibility Requirement: To be counted, a human's face must be visible within that main window.
* Real Humans Only: Ignore video game characters, cartoons, statues, or illustrated humans. Count real, physical humans only.
* Distinct Count: Accurately count each distinct, identifiable human that meets the above criteria.
* Crowd Protocol: If the main window contains a crowd (e.g., a large audience, a busy street) where individuals are too numerous or dense to count accurately, strictly return -1.

3. **Scene & Layout Detection:**
* Scene Cuts: Monitor the video for jump cuts, camera angle changes, or abrupt visual transitions to different footage. If the video is not a single, continuous, uncut shot, flag with `scene_cut: true`.
* Multiscreen/Split-Screen: Identify structural spatial divisions of the video frame. Flag as `multiscreen: true` ONLY if the screen is visibly partitioned into two or more separate, prominent video feeds (e.g., a 50/50 side-by-side interview, a multi-camera grid, or split-screen co-op). Strictly flag as `false` if there is a single main video feed, even if it contains smaller graphical overlays, Picture-in-Picture (PiP) face-cams, chat boxes, or game HUDs.

4. **Content-Transcript Resonance:**
* Analyze the semantic relationship between the *visuals* and the *transcript*, keeping in mind that the speaker is directly responding to the `Chat Bullet`.
* Assign a relevance rating based on these categories:
* "DIRECT": The video literally depicts what is being spoken about. The speaker simultaneously conducts an action that is related to their spoken response. (e.g., Bullet: "What are you cooking?", Transcript: "I am chopping onions", Video: Someone chopping onions).
* "THEMATIC": The video is related B-roll or context. The spoken response is related to the theme / environment of the video (e.g., Bullet: "How is the market?", Transcript: "The economy is crashing", Video: Background of Wall Street sign or busy traders).
* "UNRELATED": The speaker is talking about topics unrelated to what they are physically doing, often just chatting with the audience. (e.g., Bullet: "How was your day?", Transcript: "It was great, went to the park.", Video: Streamer is actively playing a first-person shooter game).
* Provide a confidence score (0.0 to 1.0) for this rating.

**Output Schema:**
You must return a single valid JSON object. Do not include markdown formatting (```json) or prose.

{
  "human_count": integer, // 0 for none, -1 for crowd/10+, count only for faces appeared in the main window.
  "is_english_transcript": bool,
  "scene_cut": bool,
  "multiscreen": bool,
  "relevance": {
    "rating": "DIRECT" | "THEMATIC" | "UNRELATED",
    "confidence": float,
    "reasoning_brief": "string (max 15 words)"
  }
}
\end{lstlisting}

\Needspace{6\baselineskip}
\subsection{Audio-Visual Captioning Prompt}

\begin{lstlisting}[style=promptstyle, caption={Prompt used for audio-visual captioning}, label={lst:prompt_caption}]
**Role**: You are an expert multimodal video analyst and dense video captioner. Your goal is to generate high-fidelity, highly descriptive training data for advanced video generation models. I will provide you with a video (or a sequence of video frames) along with its ASR (Automatic Speech Recognition) transcript.

**Task**: Accurately and vividly describe the visual content while seamlessly integrating the spoken dialogue using strict bracket notation.

**TRANSCRIPT**:
"""{TRANSCRIPT}"""

**Formatting Rule for Speech (CRITICAL)**:
Whenever you include spoken words from the transcript in your descriptions, you MUST enclose the exact spoken text within double quotes "". Do not use quotation marks for speech. The current transcript may lack punctuations, and you need to add correct punctuations according to the meaning of the sentence. You shall **add punctuations only**, and **MUST NOT modify the words**.
* *Example*: The woman smiles and says, "Welcome back to my channel."

**Generate two distinct types of descriptions based on the video provided:**

**1. Simple Description**:
Provide a concise, high-level summary that captures the environment, the primary subject, the main action, and the spoken dialogue in `""`. Use simple and accurate language to make sure easy and precise understanding. e.g.:

```
A scene from the American TV series Shameless takes place inside a family kitchen. The blonde woman spreads her hands anxiously, saying, "That man looks like he's forty years old." The man responds with a reassuring smile, "It's fine. We still have a spare room." She replies reluctantly, "Alright then. He can only stay for one night."

Two prisoners in blue uniforms lean against a wall. The Black man sighs and says reassuringly, "But that doesn't mean you're a murderer." The White man replies sorrowfully, "Someone else must have killed my wife." The Black man shakes his head with a heavy sigh.

A conversation unfolds on a British city street, framed by classic Western European-style buildings, a row of utility poles, and a British flag hanging in the background. On the left, a man in a gray suit and dark sunglasses speaks calmly, saying, "Being a gentleman has nothing to do with accent. True nobility lies in surpassing oneself."

A woman cleans a counter with a cloth in a sunlit, cozy kitchen with white cabinetry, a fridge with butterfly decals, and a vase of pink flowers. She has red hair in a bun, glasses, and casual attire, She says to camera: "did you give your neighbor his clean wood burning".
```


**2. Detailed Dense Description**:
Provide a comprehensive, continuous paragraph-length breakdown of the video. You must synthesize the visual and audio elements into a cohesive, temporally flowing narrative. Pay close attention to the following elements:

* **Subjects & Spatial Relationships**: Who or what is in the video? Describe their precise appearance, clothing (colors, textures), and where they are positioned in the frame (e.g., foreground, background, left-aligned).
* **Dialogue & Audio Synchronization**: Attribute the spoken words to the correct person on screen using the \"speech\" notation. Tie the speech directly to their visible mouth movements and body language at that specific moment.
* **Motion & Physicality**: What are the subjects doing? Describe the speed, fluidity, and physics of their actions and gestures. How do they interact with objects in their environment?
* **Setting, Lighting, & Atmosphere**: Where does the video take place? Describe the background environment, time of day, lighting setup (e.g., harsh sunlight, soft neon glow, cinematic rim lighting), and overall mood.
* **Camera Dynamics & Cinematography**: Explicitly state the camera behavior. Note if the shot is static, panning, zooming, tracking, or handheld. Mention the shot type (e.g., extreme close-up, medium shot, wide angle) and depth of field (e.g., blurred background).
* **Temporal Sequence**: Describe the video chronologically from beginning to end, ensuring the visual actions are tied to the specific lines of dialogue being spoken at that exact moment.

**Output Schema**:
You must return a single valid JSON object. Do not include markdown formatting (```json) or prose.

{
  "simple_description": "The simple description including \"spoken text\".",
  "detailed_description": "The detailed, temporally flowing dense paragraph incorporating \"spoken text\" and rich visual details."
}
\end{lstlisting}

\Needspace{6\baselineskip}
\subsection{VLM Evaluation Prompt}
\begin{lstlisting}[style=promptstyle, caption={Prompt used for VLM evaluation}, label={lst:prompt_eval}]
You are an expert evaluator judging an AI-generated audio-visual response to a user trigger.
The video you are given was produced by a generative model as a *reply* to the trigger below.

# Trigger (the comment / message the model is replying to)
\"\"\"{trigger}\"\"\"

# Optional context (may be empty)
\"\"\"{context}\"\"\"

# Your task
Watch the supplied clip carefully (both the visuals AND the audio track) and rate how well it
works as a *response* to the trigger.  Think about whether someone receiving this clip as a
reply to their comment would feel addressed.  Score every axis from 1 (terrible) to 5 (excellent).

# Axes (with anchors)
- trigger_relevance:
    1 = The clip ignores the trigger entirely / unrelated content.
    5 = Directly and unambiguously addresses what the trigger said.
- response_appropriateness:
    1 = Even if topically related, this is *not* a sensible reply (wrong reaction, wrong intent).
    5 = Clearly a fitting reaction or answer; you can see *why* it is a reply.
- tonal_match:
    1 = Tone clashes with the trigger (e.g. cheerful response to bad news).
    5 = Emotional register, formality, sarcasm/sincerity all match perfectly.
- specificity:
    1 = Generic; could plausibly reply to almost any trigger.
    5 = Picks up specific concrete entities/details from the trigger.
- av_coherence:
    1 = The audio and the video tell contradictory stories (e.g. laugh track over crying face).
    5 = Audio and visual are tightly unified and reinforce each other.
- speech_quality:
    1 = Speech is unintelligible, broken, or off-topic.
    5 = Speech is clear, well-formed, and a coherent verbal reply.
    Use null if there is no spoken speech in the clip.
- non_verbal_expressiveness:
    1 = Flat, no useful gesture/expression/SFX/music.
    5 = Non-verbal cues clearly carry response meaning.
- naturalness:
    1 = Heavy artefacts, glitches, broken motion or audio.
    5 = Clean, believable, looks like a real recorded reply.

# Additional fields
- primary_strength:  one short phrase (<=12 words) describing what the response does best.
- primary_weakness:  one short phrase (<=12 words) describing the worst flaw.
- overall:           your final 1-5 score for "how good is this as a reply".

# Output (STRICT JSON, no prose, no markdown fences, no trailing commas)
{{
  "trigger_relevance": <int 1-5>,
  "response_appropriateness": <int 1-5>,
  "tonal_match": <int 1-5>,
  "specificity": <int 1-5>,
  "av_coherence": <int 1-5>,
  "speech_quality": <int 1-5 or null>,
  "non_verbal_expressiveness": <int 1-5>,
  "naturalness": <int 1-5>,
  "primary_strength": "<string>",
  "primary_weakness": "<string>",
  "overall": <int 1-5>
}}
"""
\end{lstlisting}

\section{Generation Demos}
\label{app:demos}

\begin{figure}[!htbp]
    \centering
    \includegraphics[width=0.92\linewidth]{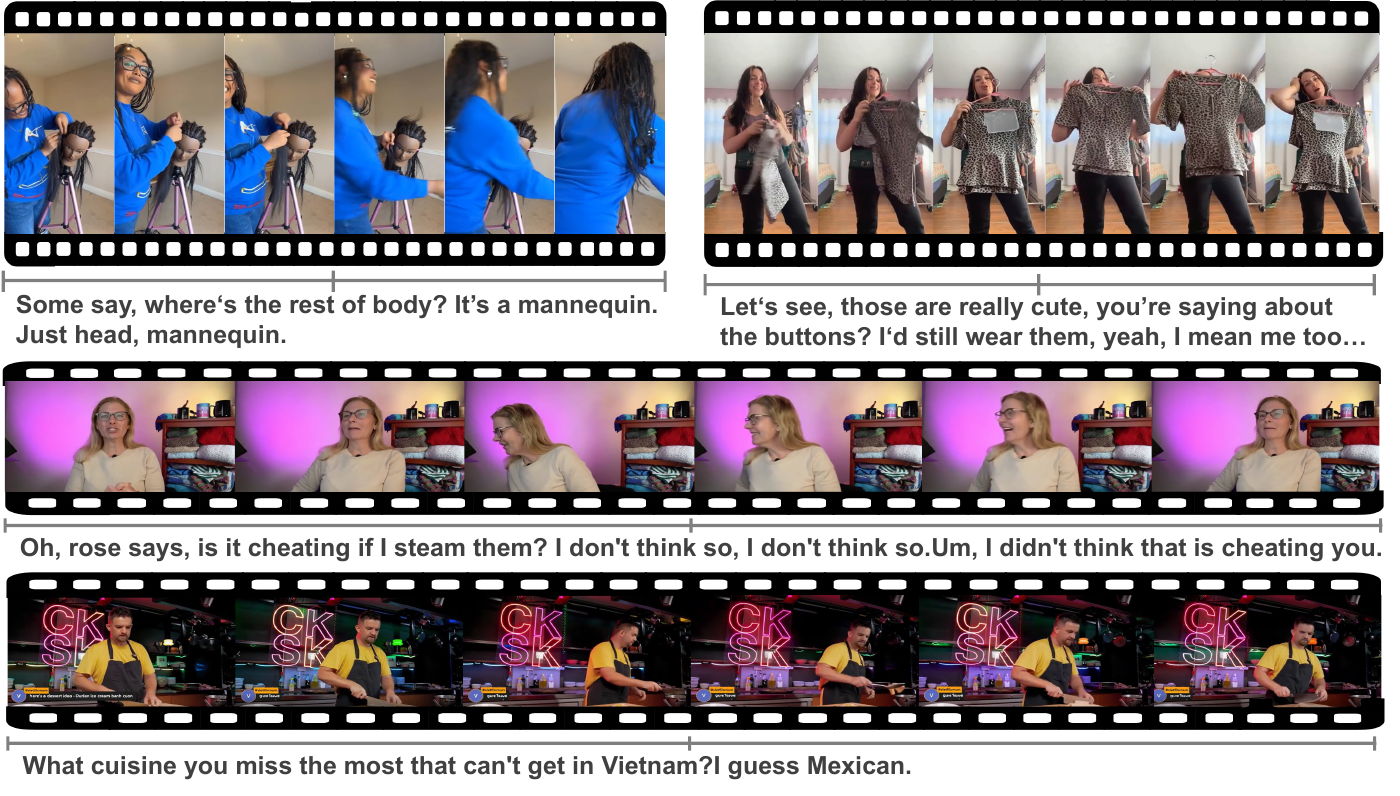}
    \caption{Generation results of the fine-tuned MOVA co-generator given the oracle ground-truth caption as conditioning text. This setting isolates rendering quality from response planning.}
    \label{fig:oracle_demo}
\end{figure}

\begin{figure}[p]
    \centering
    \includegraphics[width=0.92\linewidth,height=0.43\textheight,keepaspectratio]{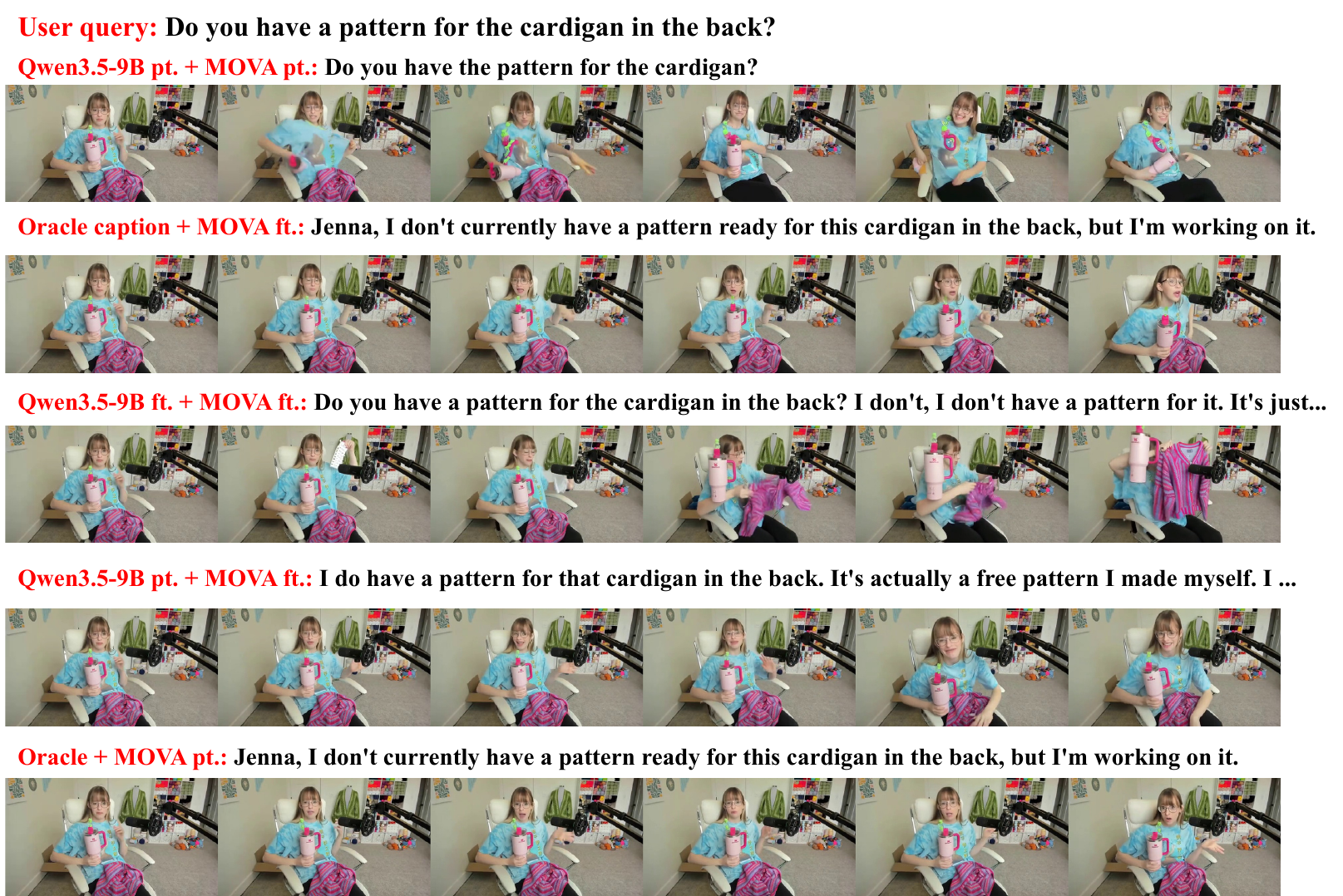}
    \caption{Qualitative comparison across pipeline configurations on a held-out interaction.}
    \label{fig:demo_compare_1}
\par\medskip
    \centering
    \includegraphics[width=0.92\linewidth,height=0.43\textheight,keepaspectratio]{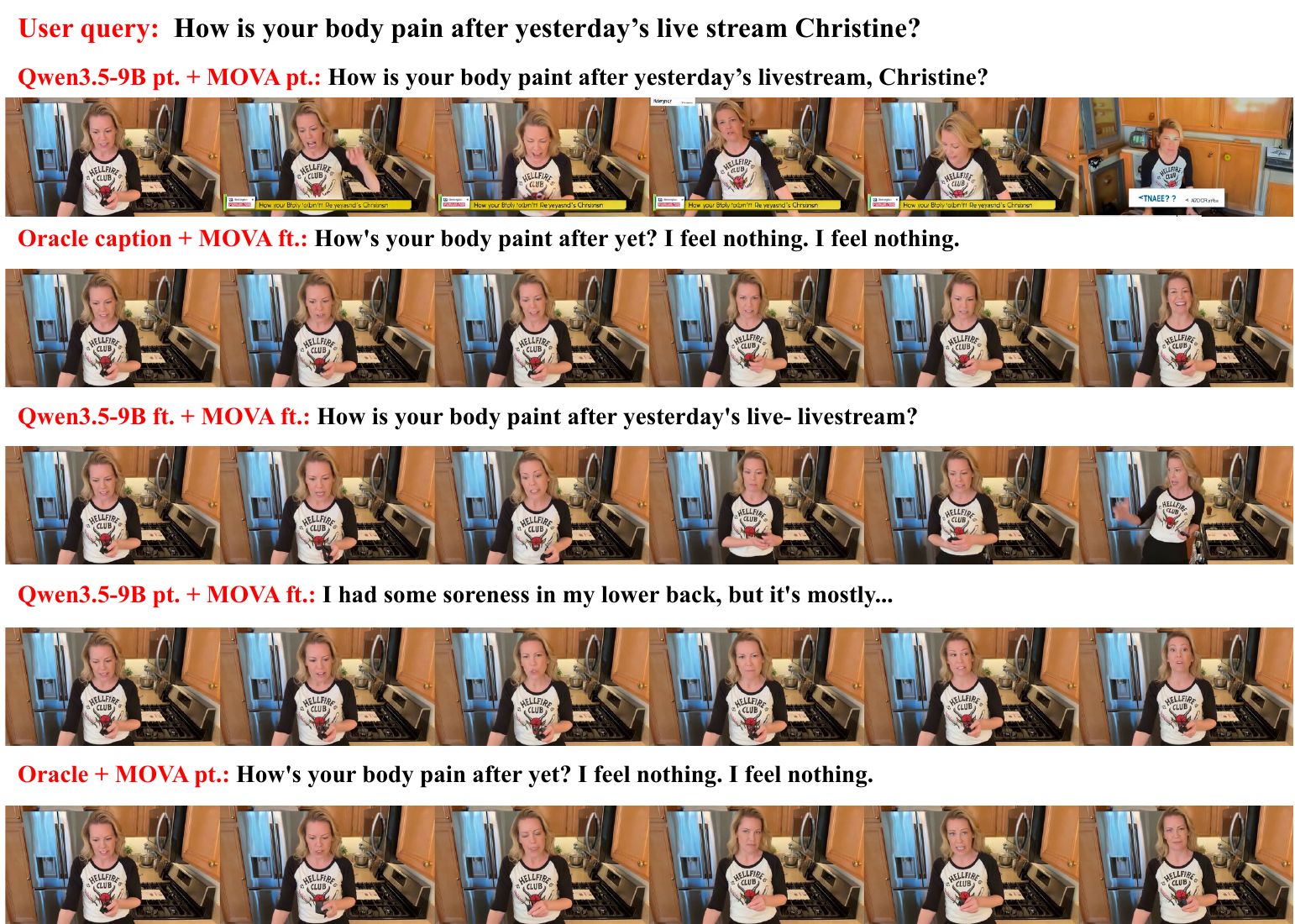}
    \caption{Qualitative comparison across pipeline configurations on a second held-out interaction.}
    \label{fig:demo_compare_2}
\end{figure}

We provide further qualitative generation results. \Cref{fig:oracle_demo} shows the audio-visual generation quality achievable given oracle ground-truth captions, which upper-bounds what the pipeline can render. Model comparisons across pipeline configurations are shown in \Cref{fig:demo_compare_1,fig:demo_compare_2}.

\noindent\textbf{Common success cases.}
The model often improves when the query refers to visible objects, simple demonstrations, or direct questions answerable from the local scene. In these cases, the generated response tends to preserve the identity and background while producing more relevant speech and motion.

\noindent\textbf{Common failure cases.}
Failure cases include long-horizon responses, ambiguous queries, interactions requiring precise hand-object manipulation, and cases where the correct response depends on information outside the visible context. Some generated videos may also exhibit audio-video desynchronization, weak lip motion, or generic gestures. These limitations suggest that future work should improve long-context modeling, response planning, and fine-grained physical control.

\section{Ethics, Licensing, and Release Considerations}
\label{app:ethics_release}

Our dataset is derived from publicly available livestream videos and associated metadata. Following standard practice for large-scale web-video corpora, and because the source videos are subject to platform terms of use and creator copyright, the release contains \emph{source identifiers and derived annotations rather than redistributed media}: video identifiers, timestamps, transcripts, captions, and the audit annotations described in \Cref{app:data_analyses}, together with preprocessing scripts that reconstruct every clip locally from the released annotations.

\noindent\textbf{Release contents.}
To make the dataset usable as an evaluation resource rather than only as a training corpus, the release includes: (i)~fixed train, validation, and genuine-query test splits, with the channel- and streamer-level disjointness described in \Cref{app:eval_set}; (ii)~explicit \texttt{query\_source} labels distinguishing real from reconstructed queries, so that either branch can be used alone; (iii)~the split-construction and leakage-prevention procedures; (iv)~preprocessing and dataset-reconstruction scripts; (v)~planner and generator training configurations; (vi)~baseline configurations together with the available adapters; and (vii)~documented failure cases.

\noindent\textbf{Privacy.}
Livestream videos may contain identifiable individuals, usernames, or personal information. We apply strict information safety controls to reduce privacy risks. Usernames and account identifiers are removed from the released metadata.

\noindent\textbf{Takedown procedure.}
We publish a privacy and takedown policy alongside the dataset and provide a removal channel. Upon a verified request, the corresponding identifiers and all derived annotations are removed from subsequent versioned releases.

\noindent\textbf{Safety filtering.}
We remove or flag samples involving unsafe, sexual, violent, or hateful content. The dataset card documents the filtering taxonomy.

\noindent\textbf{Consent and public data.}
The data originates from publicly available videos, but public availability does not by itself resolve ethical concerns about reuse. We therefore restrict the release to identifiers and derived annotations, remove account identifiers, and provide a takedown channel.

\noindent\textbf{Potential misuse.}
The dataset may improve interactive audio-visual generation, which can support creative tools, accessibility, education, and human-computer interaction. It may also be misused for impersonation, deceptive avatars, or synthetic media generation. We release the dataset with usage restrictions, a model card, and explicit prohibitions against impersonation or deceptive use.

\noindent\textbf{Distributional bias.}
The dataset reflects the distribution of publicly available livestreams and may therefore carry demographic, cultural, and platform-specific biases, which should be considered when training or evaluating downstream systems. The scope of the corpus is quantified in \Cref{sec:conclusion}.

\fi

\end{document}